\documentclass[10pt,twocolumn,letterpaper]{article}

\usepackage{cvpr}
\usepackage{times}
\usepackage{epsfig}
\usepackage{graphicx}
\usepackage{amsmath}
\usepackage{enumitem}
\usepackage{amsmath,amssymb,amsthm,amsfonts}

\usepackage{xcolor}
\usepackage{multirow}
\usepackage{caption}
\usepackage{algorithm_my}
\usepackage{algorithmic_my}
\usepackage{subfig}

\usepackage{bm}
\graphicspath{{./figures/}}

\definecolor{citecolor}{RGB}{34, 139, 34}
\definecolor{linkcolor}{rgb}{0.8, 0.12, 0.12}

\usepackage[pagebackref=true,breaklinks=true,letterpaper=true,colorlinks,bookmarks=false,citecolor=citecolor,linkcolor=linkcolor]{hyperref}

\newtheorem{lemma}{Lemma}
\newtheorem{theorem}{Theorem}

\newtheorem{corollary}{Corollary}

\cvprfinalcopy 

\def\cvprPaperID{10521} 

\ifcvprfinal\fi
\begin{document}

\title{Few Sample Knowledge Distillation for Efficient Network Compression}

 \author{Tianhong Li$^1$\thanks{This work was done when Tianhong Li was intern at Intel Labs under supervised by Jianguo Li.}\quad Jianguo Li$^2$\quad Zhuang Liu$^3$\quad  Changshui Zhang$^4$\\
 $^1$MIT\quad $^2$Intel Labs\quad $^3$UC Berkeley\quad $^4$Dept. Automation, Tsinghua University\\
 {\tt\small tianhong@mit.edu, jianguoli@intel.com, zhuangl@berkeley.edu, zsc@tsinghua.edu.cn}
 }

\maketitle
\thispagestyle{empty}

 \begin{abstract}
 Deep neural network compression techniques such as pruning and weight tensor decomposition usually require fine-tuning to recover the prediction accuracy when the compression ratio is high. However, conventional fine-tuning suffers from the requirement of a large training set and the time-consuming training procedure. This paper proposes a novel solution for knowledge distillation from \textbf{label-free few samples} to realize both data efficiency and training/processing efficiency.
 We treat the original network as ``teacher-net'' and the compressed network as ``student-net''.
 A 1$\times$1 convolution layer is added at the end of each layer block of the student-net, and we fit the block-level outputs of the student-net to the teacher-net by estimating the parameters of the added layers.
 We prove that the added layer can be merged without adding extra parameters and computation cost during inference.
 Experiments on multiple datasets and network architectures verify the method's effectiveness on student-nets obtained by various network pruning and weight decomposition methods. Our method can recover student-net's accuracy to the same level as conventional fine-tuning methods in minutes while using only 1\% label-free data of the full training data.
 \end{abstract}

 \section{Introduction}
 \begin{figure*}[]
     \centering
     \small
     \includegraphics[width=0.8\textwidth]{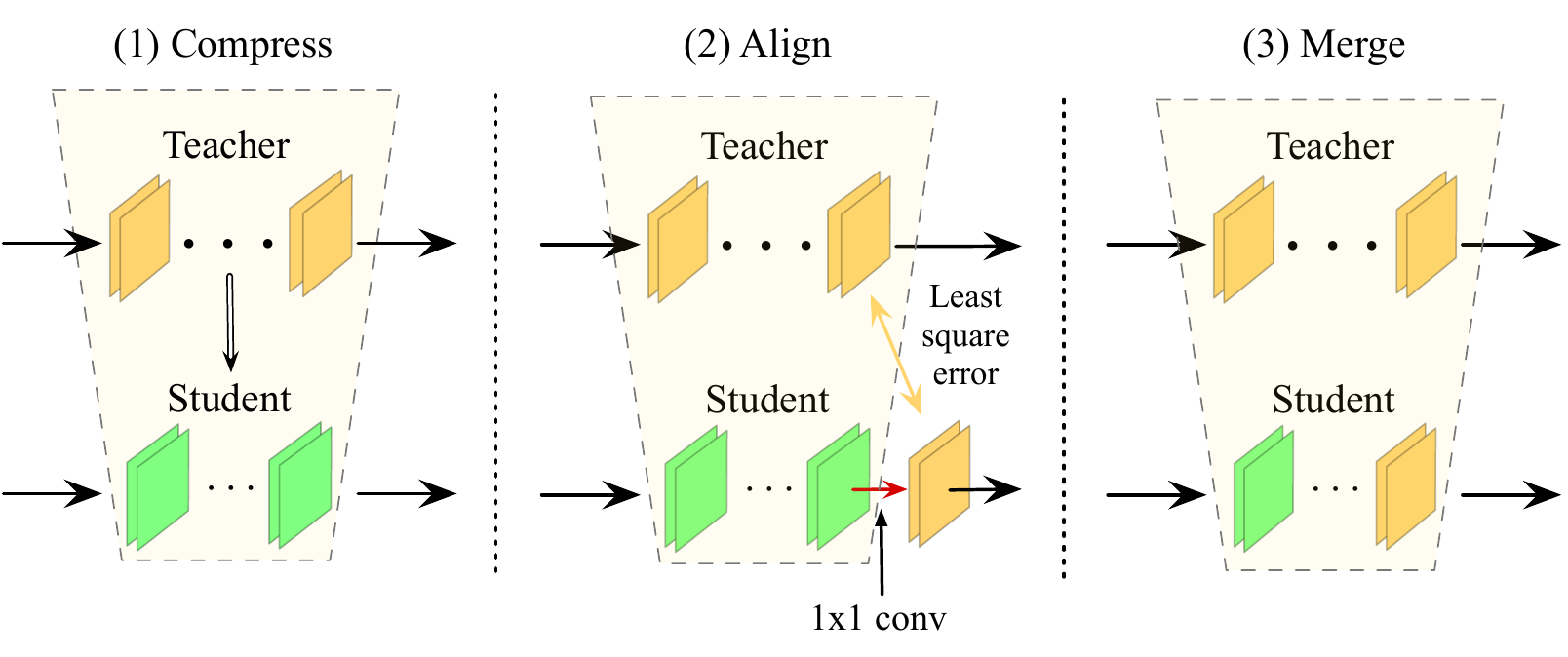}
     \vspace{-2ex}
     \caption{Three-step of few-sample knowledge distillation. (1) obtain student-net by compressing teacher-net; (2) add an 1$\times 1$ conv-layer at the end of each block of student-net, and align teacher and student by estimating the 1$\times 1$ conv-layer's parameter using least-squared regression; (3) Merge the added 1$\times$ 1 conv-layer into the previous conv-layer to obtain final student-net. }\label{fig:overview}
     \vspace{-3ex}
 \end{figure*}

 Deep neural networks have demonstrated extraordinary success in a variety of fields such as computer vision \cite{alexnet12,he2016deep}, speech recognition \cite{hinton2012deep}, and natural language processing \cite{mikolov2010recurrent}.
 However, their resource-hungry nature greatly hinders their wide deployment in some resource-limited scenarios.
 To address this limitation, many works have been done to accelerate and/or compress neural networks, among which network pruning \cite{han2015deep,li2016pruning} and network weight decomposition \cite{Denton2014Exploiting,Jaderberg2014Speeding} are particularly popular due to their competitive performance and compatibility.

 Network pruning \cite{li2016pruning,liu2017learning} and weight decomposition \cite{Zhang2016Accelerating,kim2015compression} methods can produce extremely compressed networks, but they usually suffer from significant accuracy drops so that fine-tuning is required for possible accuracy recovery. However, current fine-tuning practices are not only time-consuming but also require a fully annotated large-scale training set, which may be infeasible in practice due to privacy or confidential issues. This is \textbf{very common} when software or hardware vendors help their customer optimizing deep learning solutions. As a result, when the compression ratio is high, current methods may not recover the dropped accuracy if there are very few training examples (labeled or even unlabeled) for deployment.

 To solve this problem, one may ask if it is possible to utilize the knowledge from the original large network to the compressed compact network, since the latter has a similar block-level structure as the former, and already inherits some of the feature representation power from it. A natural solution is to use knowledge distillation (KD) \cite{bucilua2006model,ba2014deep,hinton2015distilling}, a method for transferring the knowledge from a large ``teacher'' model to a compact yet efficient ``student'' model by matching certain statistics between them. Further research introduced various kinds of matching mechanisms \cite{romero2014fitnets,srinivas2018knowledge,huang2017like,li2020gan}. The distillation procedure typically designs a loss function based on the matching mechanisms and optimizes the loss during a full training process. As a result, these methods still require time-consuming training procedure along with fully annotated large-scale training dataset, thus fail to meet our goal of training/processing efficiency and high sample-efficiency.

 This paper addresses this issue with a simple and novel method, namely few-sample knowledge distillation (FSKD), for efficient network compression, where \textit{``efficient''} here means both \textit{training/processing efficiency} and \textit{label-free data-sample efficiency}.
 As shown in \autoref{fig:overview}, FSKD contains three steps: compressing teacher-net to obtain student-net, aligning student-teacher with the added layers in student-net, and merging the added layers.
 We first obtain the student-net by pruning or decomposing the teacher-net using existing methods. In this stage, we ensure that both teacher-net and student-net have the same feature map sizes at each corresponding layer block.
 Second, we add a 1$\times$1 conv-layer at the end of each block of the student-net. We then forward the few unlabeled data to both the teacher-net and student-net, and align the block-level outputs of the student with the teacher by estimating the parameters of the added layer, using least square regressions. Because there are very few parameters to estimate in the added conv-layers, we could obtain a good estimation with a very small amount of label-free samples.
 The aligned student-net has the same number of parameters and computation cost as the original one since we prove that the added 1$\times$1 convolution can be merged into the previous convolution layer.

 FSKD has many potential usages, especially when full fine-tuning or re-training is infeasible in practice, or the data at hand is only very limited.
 We name a few concrete cases below.
 \textit{First}, edge devices have limited computing resources, while FSKD offers the possibility of on-device learning to compress deep models.
 \textit{Second}, FSKD may help cloud services obtain a compact model when only a few unlabeled data is uploaded by the customer due to privacy or confidential issues.
 \textit{Third}, FSKD enables fast model convergence if there is a strict time budget for training/fine-tuning.
 The strong practice requests have been addressed by several recent workshops \footnote{\scriptsize{http://sites.google.com/view/icml2019-on-device-compact-dnn}}.

 Our major contributions can be summarized as follows:
 \begin{itemize}
     \vspace{-1ex}
     \setlength{\topsep}{1pt}
     \setlength{\itemsep}{1pt}
     \setlength{\parskip}{1pt}
     \item To the best of our knowledge, we are the first to show that accuracy recovery from a compressed network can be done with \textbf{few unlabeled samples} within minutes using knowledge distillation on desktop PC.
     \item Extensive experiments show that the proposed FSKD method is widely applicable to deep neural networks compressed by different pruning or decomposition-based methods.
     \item We demonstrate significant accuracy improvement from FSKD over existing distillation techniques, as well as compression-ratio and speedup gain over traditional pruning/decomposition-based methods on various datasets and network architectures.
 \end{itemize}

 \section{Related Work}
 \textbf{Network Pruning} methods obtain a small network by pruning weights from a trained larger network, which can keep the accuracy of the larger model if the prune ratio is set properly. \cite{han2015learning} proposes to prune the individual weights that are near zero. Recently, filter pruning has become increasingly popular thanks to its better compatibility with off-the-shelf computing libraries, compared with weights pruning. Different criteria have been proposed to select the filters to be pruned, including norm of weights \cite{li2016pruning}, scales of multiplicative coefficients \cite{liu2017learning}, statistics of next layer \cite{luo2017thinet}, etc.
 These pruning based methods usually require iterative loop between pruning and fine-tuning for achieving better pruning ratio and speedup.
 Meanwhile, \textbf{Network Decomposition} methods try to factorize parameter-heavy layers into multiple lightweight ones.
 For instance, it may adopt low-rank decomposition to fully-connection layers \cite{Denton2014Exploiting}, and different kinds of weight decomposition to conv-layers \cite{Jaderberg2014Speeding,kim2015compression,Zhang2016Accelerating}.
 However, aggressive network pruning or network decomposition usually lead to large accuracy drops, thus fine-tuning is a must to alleviate those drops \cite{li2016pruning,liu2017learning}.

 \textbf{Knowledge Distillation} (KD) transfers knowledge from a pre-trained large teacher-net (or even an ensemble of networks) to a small student-net,
 for facilitating the deployment at test time. Originally, this is done by regressing the softmax output of the teacher model \cite{hinton2015distilling}.
 The soft continuous regression loss used here provides richer information than the label based loss,
 so that the distilled model can be more accurate than training on labeled data with cross-entropy loss. Later, various works have extended this approach by matching other statistics, including intermediate feature responses \cite{romero2014fitnets,chen2015net2net}, gradient \cite{srinivas2018knowledge}, distribution \cite{huang2017like}, Gram matrix \cite{yim2017gift}, etc. More complicatedly, deep mutual learning \cite{zhang2017dml} trains a cohort of student-nets and teaches each other collaboratively with model distillation throughout the training process. The student-nets in KD are usually designed with random weight initialization, and thus all these methods require a large amount of data (known as the ``transfer set'') to transfer the knowledge.

 As a result, it is of great interest to start the student-nets with extremely pruned or decomposed networks and explore a KD solution under the few-sample setting. The proposed FSKD has a quite different philosophy on aligning intermediate responses to the closest knowledge distillation method FitNet \cite{romero2014fitnets}.
 FitNet re-trains the whole student-net with intermediate supervision as well as label supervision using a larger amount of fully-annotated data, while FSKD only estimates parameters of the added 1$\times$1 conv-layer with few unlabeled samples. Experiments verify that FSKD is not only more efficient but also more accurate than FitNets.

 \textbf{Learning with few samples} has been extensively studied under the concept of one-shot or few-shot learning.
 One category of methods directly model few-shot samples with generative models \cite{fei2006one,lake2011one,chen2019data,bhardwaj2019dream},
 while most others study the problem under the notion of transfer learning \cite{bart2005cross,ravi2016opt}.
 In the latter category, meta-learning methods \cite{vinyals2016matching,finn2017model} solve the problem in a learning-to-learn fashion, which has been recently gaining momentum due to their application versatility.
 Most studies are devoted to the image classification task, while it is still less-explored for knowledge distillation from few samples.
 Recently, some works tried to address this problem.
 \cite{kimura2018imitation} constructs pseudo-examples using the inducing point method and develops a complicated algorithm to optimize the model and pseudo-examples alternatively.
 \cite{lopes2017data} records per-layer meta-data for the teacher-net in order to reconstruct a training set, and then adopts a standard training procedure to obtain the student-net. Both are very costly due to the complicated and heavy training procedure.
 On the contrary, we aim for an efficient solution for knowledge distillation from few unlabeled samples.

 \section{Method}
 \subsection{Overview}
 Our FSKD method consists of three steps as shown in \autoref{fig:overview}.
 \textit{First}, we obtain a student-net either by pruning or by decomposing the teacher-net.
 \textit{Second}, we add a 1$\times$1 conv-layer at the end of each block of the student-net and align the block-level outputs between teacher and student by estimating the parameters of the added layer from few unlabeled samples.
 \textit{Third}, we merge the added 1$\times$1 conv-layer into the previous conv-layer so that it will not introduce extra parameters and computation cost into the student-net.

 Three reasons make the idea works efficiently.
 \textit{First}, the compressed student-net inherits partial representation power from the teacher network, so adding 1$\times$1 conv-layer is enough to calibrate the student-net and restore the accuracy.
 \textit{Second}, the 1$\times$1 conv-layers have relatively fewer parameters, which do not require too many data for the estimation.
 \textit{Third}, the block-level output from teacher-net provides rich information as shown in FitNet \cite{romero2014fitnets}.
 Below, we will first describe our algorithm for block-level output alignment, and then prove why the added 1$\times$1 conv-layer can be merged into the previous conv-layer.

 \begin{algorithm}[tb]
     \begin{small}
         \caption{Block-coordinate descent algorithm for FSKD}
         \label{fskd}
         \begin{algorithmic}[1]
             \INPUT {Student-net $\bm{s}$, teacher-net $\bm{t}$, input data $\{\bm{X}_i\}_{i=1}^N$,\\ number of aligned blocks $M$, number of iterations $T$}
             \FOR{$k=1:T$}
             \STATE Random flip input dataset to obtain $\{\bm{X}'_i\}_{i=1}^{N}$;
             \FOR{$j=1:M$}
             \STATE Feed $\{\bm{X}'_i\}_{i=1}^N$ to the end of the $j$-th block of $\bm{t}$, obtain response $\{\bm{X}_{ij}^t\}$;
             \STATE Feed $\{\bm{X}'_i\}_{i=1}^N$ to the end of the $j$-th block of $\bm{s}$, obtain response $\{\bm{X}_{ij}^s\}$;
             \STATE Add 1$\times$1 conv-layer with tensor $\mathbf{Q}_{j}$ to the end of $j$-th block of $\bm{s}$;
             \STATE Solve $\mathbf{Q}_{j}$ with least-square regression based on \autoref{eq:Q};
             \STATE Merge $\mathbf{Q}_{j}$ into previous conv-layer $L_j$ with tensor $\mathbf{W}_j$ to obtain new tensor $\mathbf{W}'_j$ based on \autoref{th1} for student-net;
             \STATE Update the $j$-th block of $\bm{s}$, to obtain $\bm{s}'$;
             \STATE $\bm{s} = \bm{s}'$;
             \ENDFOR
             \ENDFOR
             \OUTPUT Merged conv-layers $\{\mathbf{W}'_j\}_{j=1}^M$, updated student-net $\bm{s'}$;
         \end{algorithmic}
     \end{small}
 \end{algorithm}
 \subsection{Block-level Alignment}
 In this section, we provide details of our block-level output alignment algorithm.
 Suppose $\bm{X}^s, \bm{X}^t \in \mathbb{R}^{n_o\times d}$ are the block-level output in matrix form for the student-net and teacher-net respectively, where $d$ is the per-channel feature map resolution size.
 We add a 1$\times$1 conv-layer $\mathbf{Q}$ at the end of each block of student-net before non-linear activation.
 As $\mathbf{Q}$ is degraded to the matrix form, it can be estimated with least squared regression as
 \begin{equation}\label{eq:Q}
 \small
 \mathbf{Q}^* = \arg\min_{\mathbf{Q}} \sum\nolimits_{i=1}^N \|\bm{Q}*\mathbf{X}_i^s -\bm{X}_i^t\|,
 \end{equation}
 where $N$ is the number of label-free samples used, and ``*'' means matrix product.
 The number of parameters of $\mathbf{Q}$ is $n_o \times n_o$,
 where $n_o$ is the number of output channels in the block, which is usually not too large so that we can estimate $\mathbf{Q}$ with a limited number of samples.

 Suppose there are $M$ corresponding blocks in the teacher-net and the student-net required to align, to achieve our goal, we need minimize the following loss function
 \begin{equation}\label{eq:loss}
 \small
 \mathcal{L}(\mathbf{Q}_j) = \sum\nolimits_{j=1}^M \sum\nolimits_{i=1}^N \|\mathbf{Q}_j*\bm{X}_{ij}^s -\bm{X}_{ij}^t\|_F,
 \end{equation}
 where $\mathbf{Q}_j$ is the tensor for the added 1$\times$1 conv-layer of the $j$-th block.
 In practice, we optimize this loss with a block-coordinate descent (BCD) algorithm \cite{xu2017globally}, which greedily handles each of the $M$ blocks in the student-net sequentially as shown in Algorithm-\ref{fskd} (FSKD-BCD). We can also minimize this loss using standard SGD on all added parameters (FSKD-SGD).
 However, FSKD-BCD has the following advantages over FSKD-SGD:
 (1) The BCD algorithm processes each block greedily with a sequential update rule, and each $\mathbf{Q}$ can be solved with the same set of few samples by aligning the block-level responses between teacher-net and student-net, while standard SGD considers $\{\mathbf{Q}_j\}$ all together which theoretically requires more data.
 (2) The BCD algorithm is much more efficient, which can be usually done in less than a minute.

 Unless otherwise noted, we use the FSKD-BCD algorithm with one iteration in our experiments. Appendix-A evaluates FSKD with more BCD iterations. Appendix-B makes a comparison between FSKD-BCD and FSKD-SGD.

 \subsection{Mergeable 1$\times$1 conv-layer}
 Now we prove that the added 1x1 conv-layer can be merged into the previous conv-layer without introducing additional parameters and computation cost during inference.

 \begin{theorem}\label{th1}
     A pointwise convolution with tensor $\mathbf{Q} \in \mathbb{R}^ {n'_o\times n'_i\times 1\times 1}$ can be merged into the previous convolution layer with tensor
     $\mathbf{W} \in \mathbb{R}^ {n_o\times n_i\times k\times k}$ to obtain the \textbf{merged tensor} $\mathbf{W}' = \mathbf{Q} \circ \mathbf{W}$, where $\circ$ is merging operator and $\mathbf{W}' \in \mathbb{R} ^{n'_o\times n_i\times k \times k}$
     if the following conditions  are satisfied.
     \begin{itemize}
         \vspace{-1ex}
         \setlength{\topsep}{0pt}
         \setlength{\itemsep}{1pt}
         \setlength{\parskip}{1pt}
         \item[c1.] The output channel number of $\mathbf{W}$ equals to the input channel number of $\mathbf{Q}$, i.e., $n_o = n'_i$.
         \item[c2.] No non-linear activation layer like ReLU \cite{nair2010relu} between $\mathbf{W}$ and $\mathbf{Q}$.
     \end{itemize}
 \end{theorem}
 The pointwise convolution can be viewed as a linear combination of the kernels in the previous convolution layer. Due to the space limitation, we put the formal proof and the detailed form of the merging operator in Appendix-C.
 The number of output channels of $\mathbf{W}'$ is $n'_o$, which is different from that of $\mathbf{W}$ (i.e., $n_o$).
 It is easy to have the following corollary.
 \begin{corollary}\label{th.absorb}
     When the following condition is satisfied,
     \begin{itemize}
         \item[c3.] the number of input and output channels of $\mathbf{Q}$ equals to the number of output channel of $\mathbf{W}$, i.e., $n'_i = n'_o = n_o$,
         $\mathbf{Q} \in \mathbb{R} ^{n_o\times n_o\times 1 \times 1}$,
     \end{itemize}
     the merged convolution tensor $\mathbf{W'}$ has the same parameters and computation cost as $\mathbf{W}$,
     i.e. both $\mathbf{W', W} \in \mathbb{R} ^{n_o\times n_i\times k \times k}$.
 \end{corollary}
 This condition is required not only for ensuring the same parameter size and computing cost, but also for ensuring the output-size of current layer matching to the input-size of  next layer so that these two layers are connectable.

 \section{Experiments}
 We perform extensive experiments on different image classification datasets to verify the effectiveness of FSKD on various student-net construction methods and its advantages over existing distillation methods in terms of both accuracy and speed.
 Student-nets can be obtained either by pruning based methods such as filter pruning \cite{li2016pruning} and network slimming \cite{liu2017learning},
 or by decomposition-based methods such as network decoupling \cite{guo2018nd}.
 We implement the code with PyTorch, and conduct experiments on a desktop PC with Intel i7-7700K CPU and one NVidia 1080TI GPU. The code will be made publicly available. For all
 experiments, the results are averaged over 5 trials of different randomly selected images.
 \subsection{Student-net from Pruning teacher-net}\label{sec:compress}
 \begin{figure}[]
     \centering
     \small
     \includegraphics[width=0.9\linewidth]{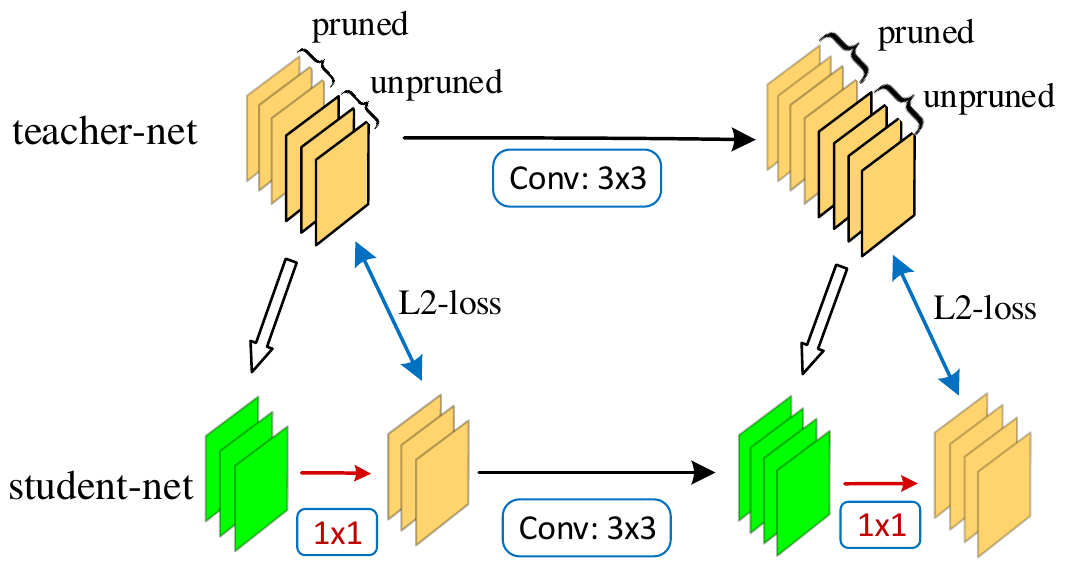}
     \vspace{-1ex}
     \caption{Illustration of FSKD on filter pruning and network slimming. At each block,
         we copy weights of the unpruned part in teacher-net to student-net, and align the feature maps of student-net to those unpruned feature maps of teacher-net by adding a 1$\times$1 conv-layer (red-color) with L2-loss. The added 1$\times$1 layer can be merged into the previous conv-layer in student-net.}\label{fig:overview_pruning}
     \vspace{-3ex}
 \end{figure}

 \subsubsection*{\textbf{Filter Pruning}}
 We first obtain the student-nets using the filter pruning method \cite{li2016pruning}, which prunes out conv-filters according to the $L_1$ norm of their weights.
 The $L_1$ norm of filter weights are sorted and the smallest portion of filters will be pruned to reduce the number of filter-channels in a conv-layer.
 \autoref{fig:overview_pruning} illustrates how FSKD works for block-level alignment in this case.
 Note that the number of channels in teacher-net may be different from that in student-net. However, we only match the \textit{un-pruned part} of feature-maps in teacher-net to the feature maps in the student-net so that FSKD is applicable in this case.
 \setlength{\tabcolsep}{4pt}
 \renewcommand{\arraystretch}{1.15}
 \begin{table}[]
     \small
     \centering
     \resizebox{0.99\linewidth}{!}{%
         \begin{tabular}{lc|cc|cc}
             \hline
             & Acc.(\%) & FLOPs($\times 10^8$) & Speedup & \#Param($\times 10^6$) & Pruned\\
             \hline
             VGG-16          & 92.66 & 3.11  & 1.00$\times$  & 15 & - \\
             Prune-A   & 85.42 & 2.06  & 1.51$\times$ & 5.3& 64\%\\
             Prune-B   & 47.90 & 1.33  & 2.34$\times$ & 3.4 & 77\%\\
             Prune-C   & 13.05 & 1.09  & 2.85$\times$ & 1.8 & 88\%\\
             \hline
         \end{tabular}
     }
     \vspace{-1ex}
     \caption{Prune-A/B/C by \textbf{filter pruning} of VGG-16 on CIFAR-10 and their accuracy, FLOPs, \#parameters, etc.}
     \label{tab:pf_scheme}
     \vspace{-2.5ex}
 \end{table}
 \begin{table}[]
     \small
     \centering
     \resizebox{0.99\linewidth}{!}{%
         \begin{tabular}{lccc}
             \hline
             & Acc. (\%) & \#Samples & Time (sec)\\
             \hline
             VGG-16     & 92.66 & 50000 &\\
             \hline
             Prune-A + FSKD   & 92.37$\pm$0.24 & 100 & 4.8\\
             Prune-A + FitNet  & 91.23$\pm$0.41 & 100 & 48.5 \\
             Prune-A + FSKD   & {92.46}$\pm$0.15 & 500& 25.5\\
             Prune-A + FitNet  & 92.13$\pm$0.35 & 500 & 139.2\\
             Prune-A + Fine-tuning & 90.25$\pm$0.67 & 500 & 40.4\\
             Prune-A + Full fine-tuning & {92.54}$\pm$0.33 & 50000 & 1059.6\\
             \hline
             Prune-B + FSKD   & 90.17$\pm$0.31 & 100 & 3.7\\
             Prune-B + FitNet  & 88.76$\pm$0.51  & 100 & 60.1\\
             Prune-B + FSKD   & {91.21}$\pm$0.23 & 500 & 19.3\\
             Prune-B + FitNet  & 90.68$\pm$0.47 & 500 &  157.1\\
             Prune-B + Fine-tuning &83.36$\pm$0.89 & 500 & 50.3\\
             Prune-B + Full fine-tuning & 91.53$\pm$0.37 & 50000 & 1753.4\\
             \hline
             Prune-C + FSKD   & 89.55$\pm$0.35 & 100 & 7.4\\
             Prune-C + FitNet  & 85.09$\pm$0.75 & 100 & 71.3\\
             Prune-C + FSKD   & {90.41}$\pm$0.31 & 500 & 33.5\\
             Prune-C + FitNet  & 88.31$\pm$0.70 & 500 & 180.3\\
             Prune-C + Fine-tuning & 78.13$\pm$0.24 & 500 & 58.7\\
             Prune-C + Full fine-tuning & 90.77$\pm$0.33 & 50000 & 2592.3\\
             \hline
         \end{tabular}
     }
     \vspace{-1ex}
     \caption{Performance comparison between FitNet, fine-tuning, FSKD by student-nets from \textbf{filter pruning} \cite{li2016pruning} of VGG-16 with pruning scheme A/B/C on CIFAR-10.
         ``Full fine-tuning'' uses full training data. }
     \label{tab:pf_result}
     \vspace{-2ex}
 \end{table}

 We make a comprehensive study of VGG-16 \cite{simonyan2014very} on CIFAR-10 dataset to evaluate the performance of FSKD along with three different pruning settings.
 \textit{First}, we follow the original pruning scheme of \cite{li2016pruning} and obtain \textbf{Prune-A}.
 \textit{Second}, we propose another more aggressive pruning scheme named \textbf{Prune-B}, which prunes 10\% more filters in the aforementioned layers, and also pruned 20\% filters for the remaining layers.
 \textit{Third}, since previous works show that one time extremely pruning may yield the pruned network unable to recovery from fine-tuning, while the iteratively pruning and fine-tuning procedure is observed effective to obtain extreme model compression \cite{han2015deep,li2016pruning,liu2017learning},
 we propose \textbf{Prune-C} which iteratively runs the pruning and FSKD procedure as described in Appendix-D for 2 iterations to achieve higher compression rate. \autoref{tab:pf_scheme} lists the accuracy, FLOPs and \#parameter information for three student-nets obtained by these pruning schemes.

 For the few-sample setting, we randomly select 100 (10 for each category) and 500 (50 for each category) images from the CIFAR-10 training set, and keep them fixed in all experiments. We evaluate 5 different randomly-selected image set and report the mean and std of accuracy.
 \autoref{tab:pf_result} lists the results of different methods of recovering a pruned network,
 including FitNet \cite{romero2014fitnets}, fine-tuning with limited data and full training data \cite{li2016pruning}.

 \begin{figure}[]
     \centering
     \small
     \subfloat{\includegraphics[width = 0.49\linewidth]{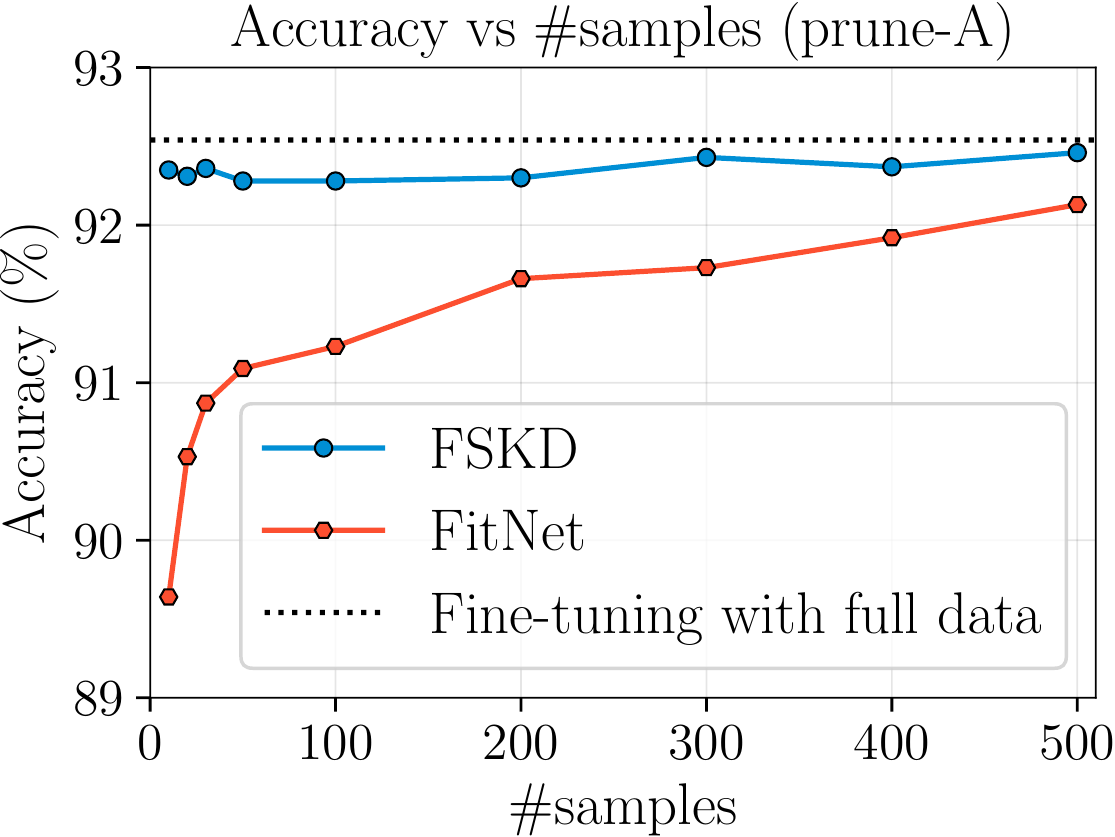}}
     \hspace{0.25ex}
     \subfloat{\includegraphics[width = 0.49\linewidth]{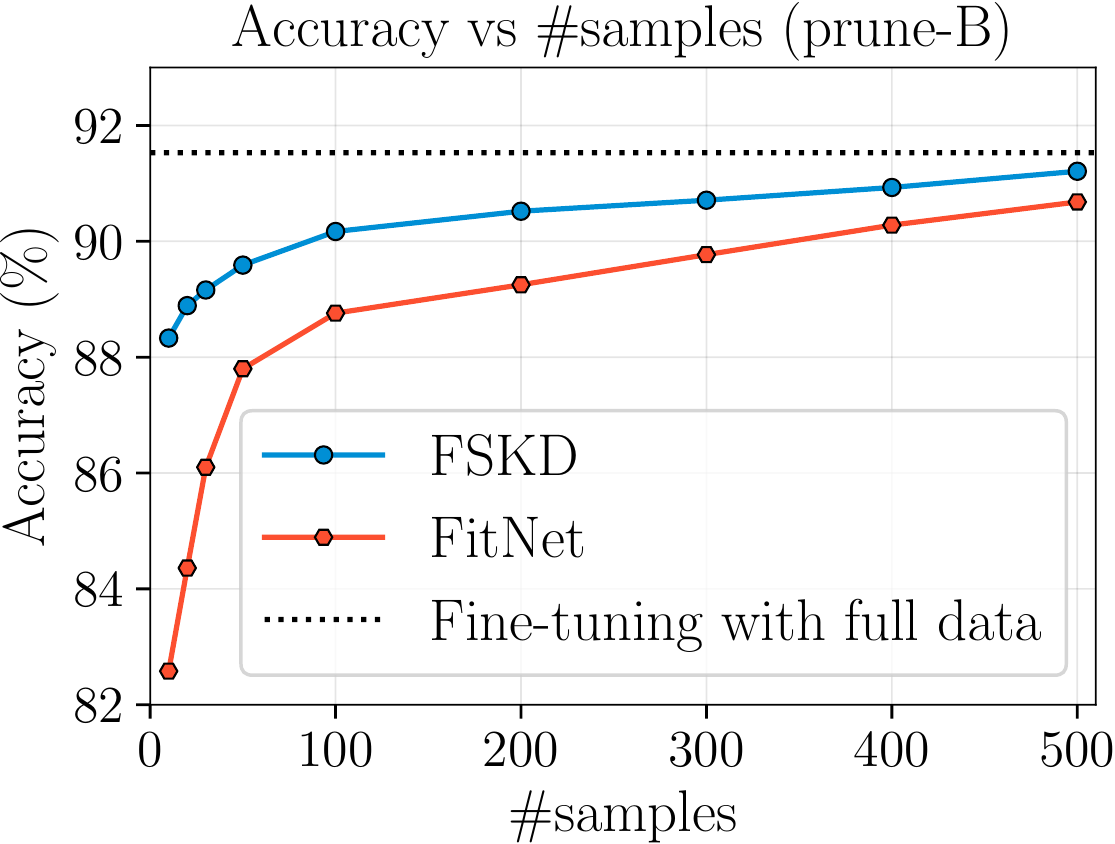}}
     \vspace{-1ex}
     \caption{Accuracy vs \#samples on CIFAR-10. Student-net Prune-A (left) Prune-B (right) by \textbf{filter pruning} \cite{li2016pruning}.}
     \label{fig:acc_data}
     \vspace{-3ex}
 \end{figure}
 \begin{figure}[]
     \centering
     \small
     \subfloat{\includegraphics[width = 0.49\linewidth]{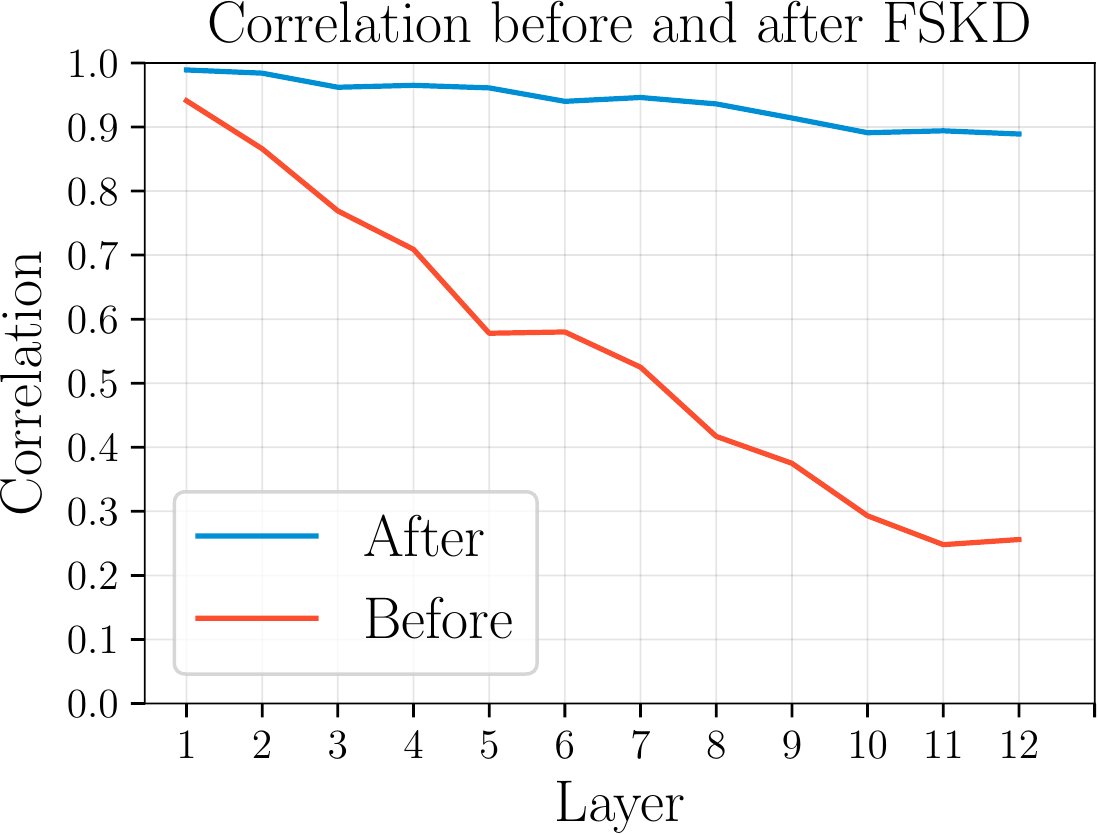}\label{fig:corr}}
     \hspace{0.25ex}
     \subfloat{\includegraphics[width = 0.49\linewidth]{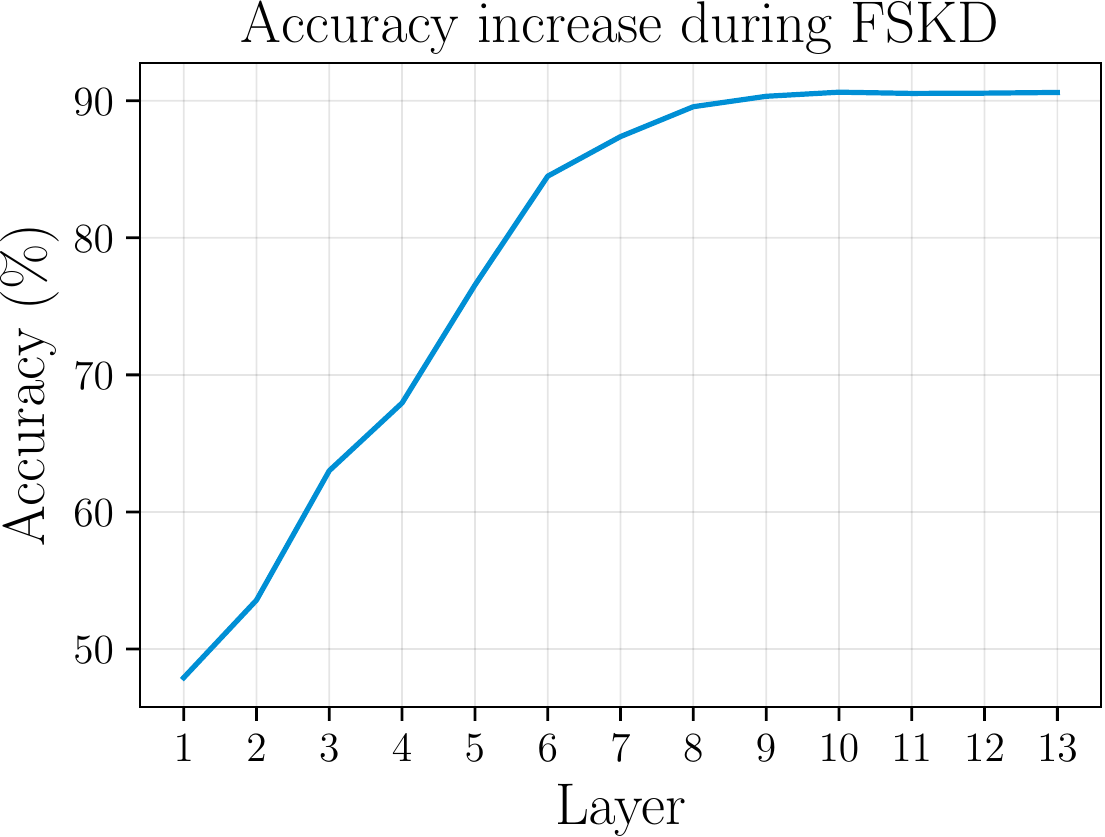}\label{fig:align}}
     \caption{Left: Layer-level output correlation between teacher-net and student-net before and after FSKD on student-nets (Prune-A) by \textbf{filter pruning} \cite{li2016pruning}. Right: Accuracy change during sequentially block-level alignment.}
     \vspace{-3ex}
 \end{figure}

 As shown in \autoref{tab:pf_result}, our method is much more efficient and provides better accuracy recovery than both FitNet and the fine-tuning procedure adopted in \cite{li2016pruning}, and also more robust with a different set of selected images. For instance, for Prune-B with only 500 samples, our method can recover the accuracy from 47.9\% to 91.2\% in 19.3s, while FitNet has to take 157.1s to recover the accuracy to 90.7\%, and few-sample fine-tuning can only recover the accuracy to 83.4\%.
 When full training set available, it takes about 30 minutes for full fine-tuning to reach similar accuracy as FSKD.
 This demonstrates the big advantages of FSKD over full fine-tuning based solutions.

 \autoref{fig:acc_data} further studies the performance versus different amount of training samples.
 Our method keeps outperforming FitNet under the same training samples.
 In particular, FitNet experiences a noticeable accuracy drop when the number of samples is less than 100, while FSKD can still recover the accuracy of the pruned network to a high level.

 \setlength{\tabcolsep}{5pt}
 \renewcommand{\arraystretch}{1.1}
 \begin{table*}[]
     \small
     \centering
     \resizebox{0.9\linewidth}{!}{%
         \begin{tabular}{lcccccccc}
             \hline
             Filter-prune-ratio \& Method & Acc. before(\%) & Acc. after(\%) & FLOPs($\times 10^8$) & Speedup & \#Param($\times 10^6$) & Pruned \\
             \hline
             VGG-19            & 93.38 & -     & 7.97  & 1.00$\times$ & 20 & -\\
             70\%  + FSKD    & 15.90 & {93.41}$\pm$0.23 & 3.91  & 2.04$\times$ & 2.2 & 89\% \\
             70\% + FitNet  & 15.90 & 90.47$\pm$0.57 & 3.91  & 2.04$\times$ & 2.2 & 89\%\\
             70\%  + Fine-tuning  & 15.90 & 62.86$\pm$2.85 & 3.91  & 2.04$\times$ & 2.2 & 89\%\\
             \hline
             ResNet-164            & 95.07 & -     & $4.99$  & 1.00$\times$  & $1.7$ & -\\
             60\%  + FSKD   & 54.46 & {94.19}$\pm$0.21 & 2.75  & 1.82$\times$ & 1.1 & 37\%\\
             60\%  + FitNet  & 54.46 & 88.94$\pm$0.66 & 2.75  & 1.82$\times$ & 1.1 & 37\%\\
             60\%  + Fine-tuning  & 54.46 & 60.94$\pm$3.12 & 2.75  & 1.82$\times$ & $1.1 $ & 37\%\\
             \hline
             DenseNet-40            & 94.18 & -     & 5.33  & 1.00$\times$  & $1.1 $ & -\\
             60\% + FSKD   & 88.24 & {93.62}$\pm$0.09 & 2.89  & 1.84$\times$ & 0.5 & 54\%\\
             60\% + FitNet  & 88.24 & 91.37$\pm$0.17 & 2.89  & 1.84$\times$ & 0.5 & 54\%\\
             60\% + Fine-tuning & 88.24 & 88.98$\pm$0.79 & 2.89  & 1.84$\times$ & 0.5 & 54\%\\
             \hline
         \end{tabular}
     }
     \vspace{-1ex}
     \caption{Performance comparison between FSKD, FitNet and fine-tuning  on different network structures obtained by \textbf{network slimming} \cite{liu2017learning} with 100 samples randomly selected from CIFAR-10 training set.}
     \label{tab:ns_10_result}
     \vspace{-1ex}
 \end{table*}
 \begin{table*}[]
     \small
     \centering
     \resizebox{0.9\linewidth}{!}{%
         \begin{tabular}{lccccccc}
             \hline
             Filter-prune-ratio \& Method & Acc. before(\%) & Acc. after(\%) & FLOPs($\times 10^8$) & Speedup & \#Param($\times 10^6$) & Pruned \\
             \hline
             VGG-19            & 72.08 & -     & 7.97  & 1.00$\times$ & 20 & - \\
             50\% + FSKD    & 9.24 & {71.98}$\pm$0.15 & 5.01  & 1.60$\times$ & 5.0 & 75\% \\
             50\% + FitNet  & 9.24 & 69.52$\pm$0.43 & 5.01  & 1.60$\times$ & 5.0 & 75\% \\
             50\% + Fine-tuning  & 9.24 & 48.75$\pm$2.86 & 5.01  & 1.60$\times$ & 5.0 & 75\% \\
             \hline
             ResNet-164            & 76.56 & -     & 5.00  & 1.00$\times$  & 1.7 & -\\
             40\% + FSKD   & 46.07 & {76.11}$\pm$0.13 & 3.33  & 1.50$\times$ & 1.5 & 14\%\\
             40\% + FitNet  & 46.07 & 73.87$\pm$0.45 & 3.33  & 1.50$\times$ & 1.5 & 14\%\\
             40\% + Fine-tuning  & 46.07 & 57.45$\pm$1.94 & 3.33  & 1.50$\times$ & 1.5 & 14\%\\
             \hline
             DenseNet-40            & 73.21 & -     & 5.33  & 1.00$\times$  & 1.1 & -\\
             40\% + FSKD   & 60.62 & {73.26}$\pm$0.07 & 3.71  & 1.44$\times$ & 0.71 & 36\%\\
             40\% + FitNet  & 60.62 & 71.08$\pm$0.33 & 3.71  & 1.44$\times$ & 0.71 & 36\%\\
             40\%  + Fine-tuning & 60.62 & 62.36$\pm$0.97 & 3.71  & 1.44$\times$ & 0.71 & 36\%\\
             \hline
         \end{tabular}
     }
     \vspace{-1ex}
     \caption{Performance comparison between FSKD, FitNet and fine-tuning on different network structures obtained by \textbf{network slimming} \cite{liu2017learning} with 500 samples randomly selected from CIFAR-100 training set.}
     \label{tab:ns_100_result}
     \vspace{-3ex}
 \end{table*}
 \begin{table*}[]
     \small
     \centering
     \resizebox{0.9\linewidth}{!}{%
         \begin{tabular}{lccccccc}
             \hline
             Filter-prune-ratio \& Method & Acc. before(\%) & Acc. after(\%) & GFLOPs & Speedup & \#Param($\times 10^6$) & Pruned \\
             \hline
             VGG-A            & 63.3 & -     & 7.74  & 1.00$\times$ & 132.9 & - \\
             50\% + FSKD    & 13.8 & 62.5$\pm$0.2 & 5.41  & 1.43$\times$ & 23.2 & 83\% \\
             50\% + FitNet  & 13.8 & 58.3$\pm$0.4 & 5.41  & 1.43$\times$ & 23.2 & 83\% \\
             50\% + Fine-tuning  & 13.8 & 19.2$\pm$3.5 & 5.41  & 1.43$\times$ & 23.2 & 83\% \\
             \hline
         \end{tabular}
     }
     \vspace{-1ex}
     \caption{Performance comparison between FSKD, FitNet and fine-tuning on VGG-A (VGG-11) network structures obtained by \textbf{network slimming} \cite{liu2017learning} with 1000 samples randomly selected from ImageNet training set.}
     \label{tab:ns_img_result}
     \vspace{-1ex}
 \end{table*}

 We further illustrate the per-layer (block) feature responses difference between teacher-net and student-net before and after using FSKD in {\color{red}Figure} \autoref{fig:corr}.
 Before applying FSKD, the correlation between teacher-net and student-net is broken due to aggressive compression.
 However, after FSKD, the per-layer correlations are mostly restored.
 This verifies the ability of FSKD for recovering lost information. We also show the accuracy change during sequentially block-level alignment in {\color{red}Figure} \autoref{fig:align}, which clearly demonstrate the effectiveness of our sequentially block-by-block update in the FSKD algorithm.

 \vspace{-1ex}
 \subsubsection*{\textbf{Network Slimming}}
 We then study the student-net from another filter pruning method named network slimming \cite{liu2017learning}, which removes insignificant filter channels and corresponding feature maps using sparsified channel scaling factors. Network slimming consists of three steps: sparse regularized training, pruning and fine-tuning.
 Here, we replace the time-consuming fine-tuning step with our FSKD,
 and follow the original paper \cite{liu2017learning} to
 conduct experiments to prune different networks on different datasets.
 The alignment framework is the same as the filter pruning case as shown in \autoref{fig:overview_pruning}.

 We apply FSKD on networks pruned from VGG-19, ResNet-164, and DenseNet-40 \cite{densenet}, on both CIFAR-10 and CIFAR-100 datasets.
 \autoref{tab:ns_10_result} lists results on CIFAR-10, while \autoref{tab:ns_100_result} lists results on CIFAR-100.
 Note that the filter-prune-ratio (like 70\% in \autoref{tab:ns_10_result}) means the portion of filters that are removed in comparison to the total number of filters in the network.
 We also apply FSKD on networks pruned from VGG-A (or VGG-11) on ImageNet dataset, as shown in \autoref{tab:ns_img_result}.

 The results show that that FSKD consistently outperforms FitNet and fine-tuning with a notable margin under the few-sample setting on all evaluated networks and datasets. This study demonstrates that FSKD is universally applicable to various network structure and pruning methods, and can recover the accuracy of the pruned network using few unlabeled samples to the same level of fine-tuning using fully annotated training dataset.
 \begin{table*}[]
     \small
     \centering
     \resizebox{0.9\linewidth}{!}{%
         \begin{tabular}{lcccccc}
             \hline
             & Acc. before(\%) & Acc. after (\%) & GFLOPs & Speedup & \#Param$^*$($\times 10^6$) & Pruned\\
             \hline
             VGG-16 (teacher)      & 68.4 & - & 15.47 & 1.00$\times$ &14.71 &-\\
             Decoupled ($T=2$) + FSKD    & 0.24 & 62.7$\pm$0.2 & 3.76  & 4.11$\times$ &3.35 &  77.2\% \\
             Decoupled ($T=3$) + FSKD    & 1.57 & 67.1$\pm$0.1 & 5.54  & 2.79$\times$ &5.02 &  65.8\% \\
             Decoupled ($T=4$) + FSKD    & 54.6 & 67.6$\pm$0.1 & 7.31  & 2.12$\times$ &6.69 &  54.5\%\\
             \hline
             ResNet-18 (teacher)    & 67.1 & - & 1.83  & 1.00$\times$ &11.17 &-\\
             Decoupled ($T=2$) + FSKD    & 0.21 & 49.5$\pm$0.5 & 0.55  & 3.33$\times$ &2.69 & 75.9\% \\
             Decoupled ($T=3$)  + FSKD   & 3.99 & 61.9$\pm$0.3 & 0.75  & 2.44$\times$ & 3.95 & 64.6\% \\
             Decoupled ($T=4$) + FSKD     & 26.5 & 65.1$\pm$0.1 & 0.95  & 1.92$\times$ & 5.20 & 53.4\% \\
             Decoupled ($T=5$) + FSKD    & 53.6 & 66.3$\pm$0.1 & 1.15  & 1.60$\times$ & 6.46 & 42.2\% \\
             \hline
         \end{tabular}
     }
     \vspace{-1ex}
     \caption{Performance of FSKD on different student-nets obtained by \textbf{network decoupling} \cite{guo2018nd} VGG-16 and ResNet-18 with different parameters $T$ on \textbf{ImageNet} dataset.
         ``$*$'' here means that parameters from FC-layer are not counted, only those from conv-layers are counted, since decoupling only handles the conv-layers. }
     \label{tab:nd}
     \vspace{-3ex}
 \end{table*}
 \begin{figure}[]
     \centering
     \small
     \includegraphics[width=0.95\linewidth]{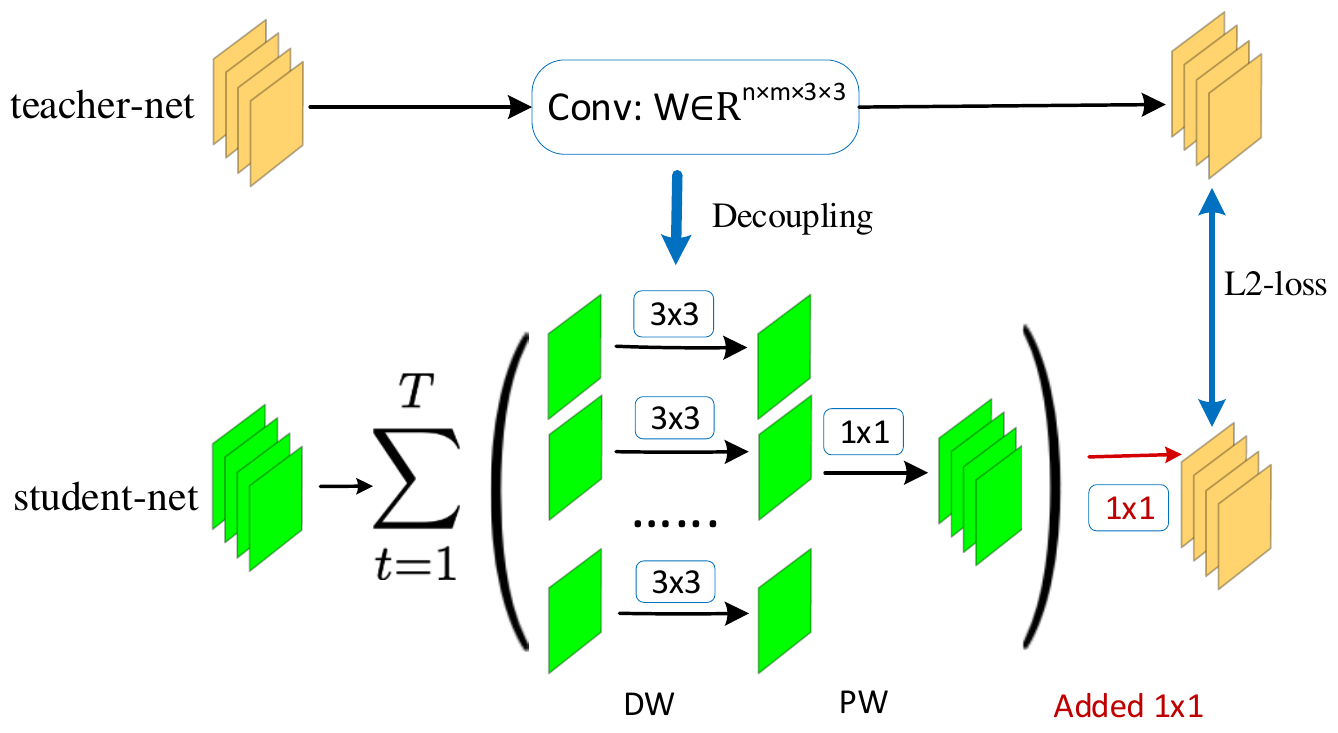}
     \vspace{-1.5ex}
     \caption{Illustration of FSKD on network decoupling. At each block of teacher-net, we decouple regular-conv into a sum of depthwise + pointwise conv-layers as the block of student-net, and align the feature maps of student-net to that of teacher-net by adding a 1$\times$1 conv-layer (red-color) with L2-loss.
         The added layer can be merged into previous the pointwise layer in student-net.}\label{fig:overview_decoupling}
     \vspace{-3ex}
 \end{figure}

 \subsection{Student-net from Decomposing teacher-net}\label{sec:decomp}
 In this section, we apply FSKD on a decomposition-based method called network decoupling \cite{guo2018nd}, which can decompose a regular convolution layer into the sum of several \textit{depth-wise separable blocks}, where each such block consists of a depth-wise (DW) conv-layer and a point-wise (PW, 1$\times$1) conv-layer. The compression ratio increases as the number ($T$) of such blocks decreases, but the accuracy of the compressed model will also drop. Since each decoupled block ends with a 1$\times$1 convolution, we can apply FSKD at the end of each decoupled block. \autoref{fig:overview_decoupling} illustrates how FSKD works for block-level alignment in this case.

 Following \cite{guo2018nd}, we obtain student-nets by decoupling VGG-16 and ResNet-18 pre-trained on ImageNet with different $T$ values.
 We evaluate the resulted network performance on the validation set of the ImageNet classification task.
 We randomly select one image from each of the 1000 classes in ImageNet training set to obtain 1000 samples as our FSKD training set.
 \autoref{tab:nd} shows the top-1 accuracy of student-net before and after applying FSKD on VGG-16 and ResNet-18.

 It is quite interesting to see that when $T$ is small, we can recover the accuracy of student-net from nearly random guess (0.24\%, 0.21\%) to a much higher level (62.7\% and 49.5\%) with only 1000 samples.
 One possible explanation is that the highly-compressed networks still inherit some representation power from the teacher-net
 i.e., the depth-wise 3$\times$3 convolution, while lacking the ability to output meaningful predictions due to
 the degraded and inaccurate 1$\times$1 convolution.
 The FSKD calibrates the 1$\times$1 convolution by aligning the block-level responses between teacher-net and student-net
 so that the lost information in 1$\times$1 convolution is compensated, and reasonable recovery is achieved.

 In all the other cases, FSKD can recover the accuracy of a highly-compressed network to be comparable with the original network. This shows that FSKD can be applied to student-net compressed by network decomposition, and that FSKD can achieve great performance on large and difficult dataset such as ImageNet.


 \section{Analysis and Discussion}
 \subsubsection*{FSKD with Arbitrary Data}
 In this section, we try to answer the following question: is FSKD totally label-free? For example, is FSKD still valid if the available few samples are arbitrary images and the teacher network never sees these images before? To answer this question, we evaluate FSKD's performance on VGG-16 model trained on CIFAR-10 and compressed using filter pruning (prune-B), with the few samples for FSKD are randomly selected from CIFAR-100 instead of CIFAR-10.

 \begin{table}[]
     \small
     \centering
     {%
         \begin{tabular}{lccc}
             \hline
             & Acc. (\%) & \#Samples\\
             \hline
             VGG-16     & 92.66 & 50000 (CIFAR-10) &\\
             \hline
             Prune-B + FSKD  & 90.17$\pm$0.31 & 100 (CIFAR-10)\\
             Prune-B + FSKD & 90.15$\pm$0.31  & 100 (CIFAR-100) \\
             Prune-B + FSKD & 91.21$\pm$0.23 & 500 (CIFAR-10)\\
             Prune-B + FSKD & 91.20$\pm$0.25 & 500 (CIFAR-100)\\
             \hline
         \end{tabular}
     }
     \vspace{-1ex}
     \caption{Performance comparison between FSKD using samples from CIFAR-10 and CIFAR-100. Student-nets from \textbf{filter pruning} \cite{li2016pruning} of VGG-16 with pruning scheme B on CIFAR-10.}
     \label{tab:cifar10_cifar100}
     \vspace{-2ex}
 \end{table}

 As shown in \autoref{tab:cifar10_cifar100}, there is no statistical difference in accuracy between FSKD using data from CIFAR-10 or CIFAR-100. This shows that FSKD aligns the student-net with the teacher-net without any information about the labels of the data. Even if the input images are of classes it has never seen before (CIFAR-100 does not include classes in CIFAR-10), FSKD can still recover the student network to the same accuracy level. This further demonstrates FSKD's potential in situations where only a few samples of unlabeled data are available.

 \vspace{-1.5ex}
 \subsubsection*{What if student-nets are  hand-designed?}
  Our previous experiments construct student-nets by pruning or decomposing teacher-net, and then apply FSKD to boost their performance.
  People may be interested in the problem ``what if student-nets are hand-designed with random initialization''.
  In fact, there are two existing works \cite{kimura2018imitation, lopes2017data} making some pioneer trials on this topic with specific methods for ``pseudo'' examples generation. Here we conduct experiments to compare our method to these two methods under the same few-sample setting on the same dataset MNIST, for a fair comparison.

  Due to different network structures used in these two methods, we make a separate comparison.
  For \cite{kimura2018imitation}, the teacher-net has 3 conv-layers followed by 2 fully-connected layers.
  For \cite{lopes2017data}, the teacher-net is a standard LeNet-5.
  For both cases, the student-net is the ``half-sized'' to that of the corresponding teacher-net in terms of the number of feature map channels per conv-layers. As the channel number between student-net and teacher-net is different, we adopt the same strategy as in \autoref{fig:overview_pruning} for filter pruning. That means, the student-net only corresponds to the un-pruned part of the teacher-net, which is obtained the same as \cite{li2016pruning}. One difference is that we did not copy the weight from un-pruned part of teacher-net to the student-net, while keeping the weight of student-net randomly initialized.
  For both cases, we compared our FSKD with (1) standard SGD trained on few samples with labeled loss; (2) method from \cite{kimura2018imitation} or \cite{lopes2017data} under the same setting;
  (3) the FitNet method trained on few samples.
  In order to better simulate the few-sample setting, we do not apply data augmentation to the training set.
  We randomly pick 10, 20, 50, 100 and 200 samples from the MNIST training set and keep these few-sample sets fixed across this study.
  \autoref{tab:zero} lists the comparison results.
  It shows that SGD with few samples performs the worst, while \cite{kimura2018imitation} performs better on the same settings than SGD (still worse in the case of 200 samples). The data-free method \cite{lopes2017data} performs better than SGD.
  On both cases, FitNet shows much better performance than SGD and the two compared methods, while our FSKD further outperform FitNet with a noticeable gap, where the gap becomes smaller and smaller when the number of samples increases.
  This may be due to the following reason.
  FSKD can be viewed as a special case of FitNet.
  FitNet optimizes all the weights between teacher-net and student-net using standard SGD algorithm, while FSKD optimizes only the weights from added 1$\times$1 conv-layers in student-net with the BCD algorithm. The BCD algorithm is more sample-efficient than the SGD based algorithm so that FSKD performs both much more efficient and accurate than FitNet on few-sample settings. When training samples used are increased, FSKD will converge to FitNet in the end.

  \setlength{\tabcolsep}{5pt}
  \renewcommand{\arraystretch}{1.1}
  \begin{table}[]
      \large
      \centering
      \resizebox{1.0\linewidth}{!}{%
          \begin{tabular}{lcccccc}
              \hline
              \#labeled data  & 10 & 20 & 50 & 100 & 200 & all-meta-data\\
              \hline
              SGD      & 37.91 & 46.0 & 66.0 & 78.3 & 86.7 &-\\
              \cite{kimura2018imitation} & 44.1 & 53.9 & 70.4  & 80.0 & 86.6&- \\
              FitNet & 86.1 & 92.3 & 94.5  & 96.0 & 96.5&-  \\
              FSKD    & \textbf{94.4} & \textbf{96.5} & \textbf{97.0}  & \textbf{97.5} & \textbf{97.8}&- \\
              \hline
              \hline
              SGD      & 57.1 & 68.3 & 81.3 & 85.8 & 89.7&- \\
              \cite{lopes2017data} & - & - & -  & - & - &92.5 \\
              FitNet & 90.3 & 94.2 & 96.1  & 96.7 & 97.3&- \\
              FSKD    & \textbf{95.5} & \textbf{97.2} & \textbf{97.6}  & \textbf{98.0} & \textbf{98.1}&- \\
              \hline
          \end{tabular}
      }
      \vspace{-1ex}
      \caption{Performance of FSKD on hand designed student-nets with random initialization, compared with previous works \cite{kimura2018imitation, lopes2017data}.}
      \label{tab:zero}
      \vspace{-2ex}
  \end{table}

 \section{Conclusion}
 We proposed a novel yet simple method, namely few-sample knowledge distillation (FSKD) for efficient network compression, while \textit{``efficient''} lies in both \textit{training/processing efficiency} and \textit{label-free  sample efficiency}.
 FSKD works for student-nets constructed by either pruning or decomposing teacher-nets with different methods. It demonstrates great efficiency over fine-tuning based solution and advantages over traditional knowledge distillation methods like FitNet by a large margin in the few-sample setting,
 with extra merits that FSKD is totally label-free in the optimization.

 \vspace{1ex}
 \noindent\textbf{Acknowledgements}: Changshui Zhang is funded by NSFC under Grant No. 61876095 \& 61751308, and Beijing Academy of Artificial Intelligence (BAAI).

 {
 \small
 \bibliographystyle{ieee_fullname}
 \bibliography{main}

\begin{thebibliography}{10}\itemsep=-1pt

\bibitem{ba2014deep}
Jimmy Ba and Rich Caruana.
\newblock Do deep nets really need to be deep?
\newblock In {\em NIPS}, 2014.

\bibitem{bart2005cross}
Evgeniy Bart and Shimon Ullman.
\newblock Cross-generalization: Learning novel classes from a single example by
  feature replacement.
\newblock In {\em CVPR}. IEEE, 2005.

\bibitem{bhardwaj2019dream}
Kartikeya Bhardwaj, Naveen Suda, and Radu Marculescu.
\newblock Dream distillation: A data-independent model compression framework.
\newblock {\em arXiv preprint arXiv:1905.07072}, 2019.

\bibitem{bucilua2006model}
Cristian Bucila, Rich Caruana, Alexandru Niculescu-Mizil, et~al.
\newblock Model compression.
\newblock In {\em SIGKDD}. ACM, 2006.

\bibitem{chen2019data}
Hanting Chen, Yunhe Wang, Chang Xu, Zhaohui Yang, Chuanjian Liu, Boxin Shi,
  Chunjing Xu, Chao Xu, and Qi Tian.
\newblock Data-free learning of student networks.
\newblock In {\em Proceedings of the IEEE International Conference on Computer
  Vision}, pages 3514--3522, 2019.

\bibitem{chen2015net2net}
Tianqi Chen, Ian Goodfellow, Jonathon Shlens, et~al.
\newblock Net2net: Accelerating learning via knowledge transfer.
\newblock In {\em ICLR}, 2016.

\bibitem{Denton2014Exploiting}
Emily Denton, Zaremba, Yann Lecun, et~al.
\newblock Exploiting linear structure within convolutional networks for
  efficient evaluation.
\newblock In {\em NIPS}, 2014.

\bibitem{fei2006one}
Li Fei-Fei, Rob Fergus, Pietro Perona, et~al.
\newblock One-shot learning of object categories.
\newblock {\em IEEE Trans PAMI}, 2006.

\bibitem{finn2017model}
Chelsea Finn, Pieter Abbeel, Sergey Levine, et~al.
\newblock Model-agnostic meta-learning for fast adaptation of deep networks.
\newblock In {\em ICML}, 2017.

\bibitem{guo2018nd}
Jianbo Guo, Yuxi Li, Weiyao Lin, Yurong Chen, and Jianguo Li.
\newblock Network decoupling: From regular to depthwise separable convolutions.
\newblock In {\em BMVC}, 2018.

\bibitem{han2015deep}
Song Han, Huizi Mao, Bill Dally, et~al.
\newblock Deep compression: Compressing deep neural networks with pruning,
  trained quantization and huffman coding.
\newblock In {\em NIPS}, 2016.

\bibitem{han2015learning}
Song Han, Jeff Pool, John Tran, William Dally, et~al.
\newblock Learning both weights and connections for efficient neural network.
\newblock In {\em NIPS}, 2015.

\bibitem{he2016deep}
K. He, X. Zhang, J. Sun, et~al.
\newblock Deep residual learning for image recognition.
\newblock In {\em CVPR}, 2016.

\bibitem{hinton2012deep}
Geoffrey Hinton, Li Deng, Dong Yu, et~al.
\newblock Deep neural networks for acoustic modeling in speech recognition: The
  shared views of four research groups.
\newblock {\em IEEE Signal Processing Magazine}, 29(6), 2012.

\bibitem{hinton2015distilling}
G. Hinton, O. Vinyals, Jeff Dean, et~al.
\newblock Distilling the knowledge in a neural network.
\newblock {\em arXiv preprint arXiv:1503.02531}, 2015.

\bibitem{hong2017iteration}
Mingyi Hong, Xiangfeng Wang, Meisam Razaviyayn, and Zhi-Quan Luo.
\newblock Iteration complexity analysis of block coordinate descent methods.
\newblock {\em Mathematical Programming}, 163(1-2), 2017.

\bibitem{densenet}
Gao Huang, Zhuang Liu, Kilian~Q Weinberger, and Laurens van~der Maaten.
\newblock Densely connected convolutional networks.
\newblock In {\em CVPR}, 2017.

\bibitem{huang2017like}
Zehao Huang and Naiyan Wang.
\newblock Like what you like: Knowledge distill via neuron selectivity
  transfer.
\newblock {\em arXiv preprint arXiv:1707.01219}, 2017.

\bibitem{Jaderberg2014Speeding}
M. Jaderberg, A. Vedaldi, A. Zisserman, et~al.
\newblock Speeding up convolutional neural networks with low rank expansions.
\newblock In {\em BMVC}, 2014.

\bibitem{kim2015compression}
Y. Kim, E. Park, S. Yoo, et~al.
\newblock Compression of deep convolutional neural networks for fast and low
  power mobile applications.
\newblock In {\em ICLR}, 2016.

\bibitem{kimura2018imitation}
Akisato Kimura, Zoubin Ghahramani, Koh Takeuchi, et~al.
\newblock Few-shot learning of neural networks from scratch by pseudo example
  optimization.
\newblock In {\em BMVC}, 2018.

\bibitem{alexnet12}
A. Krizhevsky and G. Hinton.
\newblock Imagenet classification with deep convolutional neural networks.
\newblock In {\em NIPS}, 2012.

\bibitem{lake2011one}
Brenden Lake, Ruslan Salakhutdinov, Jason Gross, et~al.
\newblock One shot learning of simple visual concepts.
\newblock In {\em Proceedings of the Annual Meeting of the Cognitive Science
  Society}, volume~33, 2011.

\bibitem{li2016pruning}
Hao Li, Asim Kadav, Durdanovic I, et~al.
\newblock Pruning filters for efficient convnets.
\newblock {\em ICLR}, 2017.

\bibitem{li2020gan}
Muyang Li, Ji Lin, Yaoyao Ding, Zhijian Liu, Jun-Yan Zhu, and Song Han.
\newblock Gan compression: Efficient architectures for interactive conditional
  gans.
\newblock {\em arXiv preprint arXiv:2003.08936}, 2020.

\bibitem{liu2017learning}
Zhuang Liu, Jianguo Li, Zhiqiang Shen, et~al.
\newblock Learning efficient convolutional networks through network slimming.
\newblock In {\em ICCV}, 2017.

\bibitem{lopes2017data}
Raphael~Gontijo Lopes, Stefano Fenu, Thad Starner, et~al.
\newblock Data-free knowledge distillation for deep neural networks.
\newblock {\em arXiv preprint arXiv:1710.07535}, 2017.

\bibitem{luo2017thinet}
J. Luo, J. Wu, W. Lin, et~al.
\newblock Thinet: A filter level pruning method for deep neural network
  compression.
\newblock In {\em ICCV}, 2017.

\bibitem{mikolov2010recurrent}
Tom{\'a}{\v{s}} Mikolov, Martin Karafi{\'a}t, Luk{\'a}{\v{s}} Burget, et~al.
\newblock Recurrent neural network based language model.
\newblock In {\em INTERSPECH}, 2010.

\bibitem{nair2010relu}
Vinod Nair and Geoffrey Hinton.
\newblock Rectified linear units improve restricted boltzmann machines.
\newblock In {\em ICML}, 2010.

\bibitem{ravi2016opt}
Sachin Ravi and Hugo Larochelle.
\newblock Optimization as a model for few-shot learning.
\newblock In {\em ICLR}, 2017.

\bibitem{romero2014fitnets}
Adriana Romero, Nicolas Ballas, Samira~Ebrahimi Kahou, et~al.
\newblock Fitnets: Hints for thin deep nets.
\newblock In {\em ICLR}, 2015.

\bibitem{simonyan2014very}
Karen Simonyan and Andrew Zisserman.
\newblock Very deep convolutional networks for large-scale image recognition.
\newblock In {\em ICLR}, 2015.

\bibitem{srinivas2018knowledge}
Suraj Srinivas and Francois Fleuret.
\newblock Knowledge transfer with jacobian matching.
\newblock {\em arXiv preprint arXiv:1803.00443}, 2018.

\bibitem{vinyals2016matching}
Oriol Vinyals, Charles Blundell, Tim Lillicrap, et~al.
\newblock Matching networks for one shot learning.
\newblock In {\em NIPS}, 2016.

\bibitem{xu2017globally}
Yangyang Xu and Wotao Yin.
\newblock A globally convergent algorithm for nonconvex optimization based on
  block coordinate update.
\newblock {\em Journal of Scientific Computing}, 72(2):700--734, 2017.

\bibitem{yim2017gift}
Junho Yim, Donggyu Joo, Jihoon Bae, et~al.
\newblock A gift from knowledge distillation: Fast optimization, network
  minimization and transfer learning.
\newblock In {\em CVPR}, 2017.

\bibitem{Zhang2016Accelerating}
X. Zhang, J. Zou, J. Sun, et~al.
\newblock Accelerating very deep convolutional networks for classification and
  detection.
\newblock {\em IEEE TPAMI}, 38(10), 2016.

\bibitem{zhang2017dml}
Ying Zhang, Tao Xiang, Timothy Hospedales, et~al.
\newblock Deep mutual learning.
\newblock In {\em CVPR}, 2018.

\end{thebibliography}
 }

\clearpage
\section*{A: FSKD with different \# BCD iterations}

\begin{figure}[h]
\small
\centering
\includegraphics[height=0.6\linewidth]{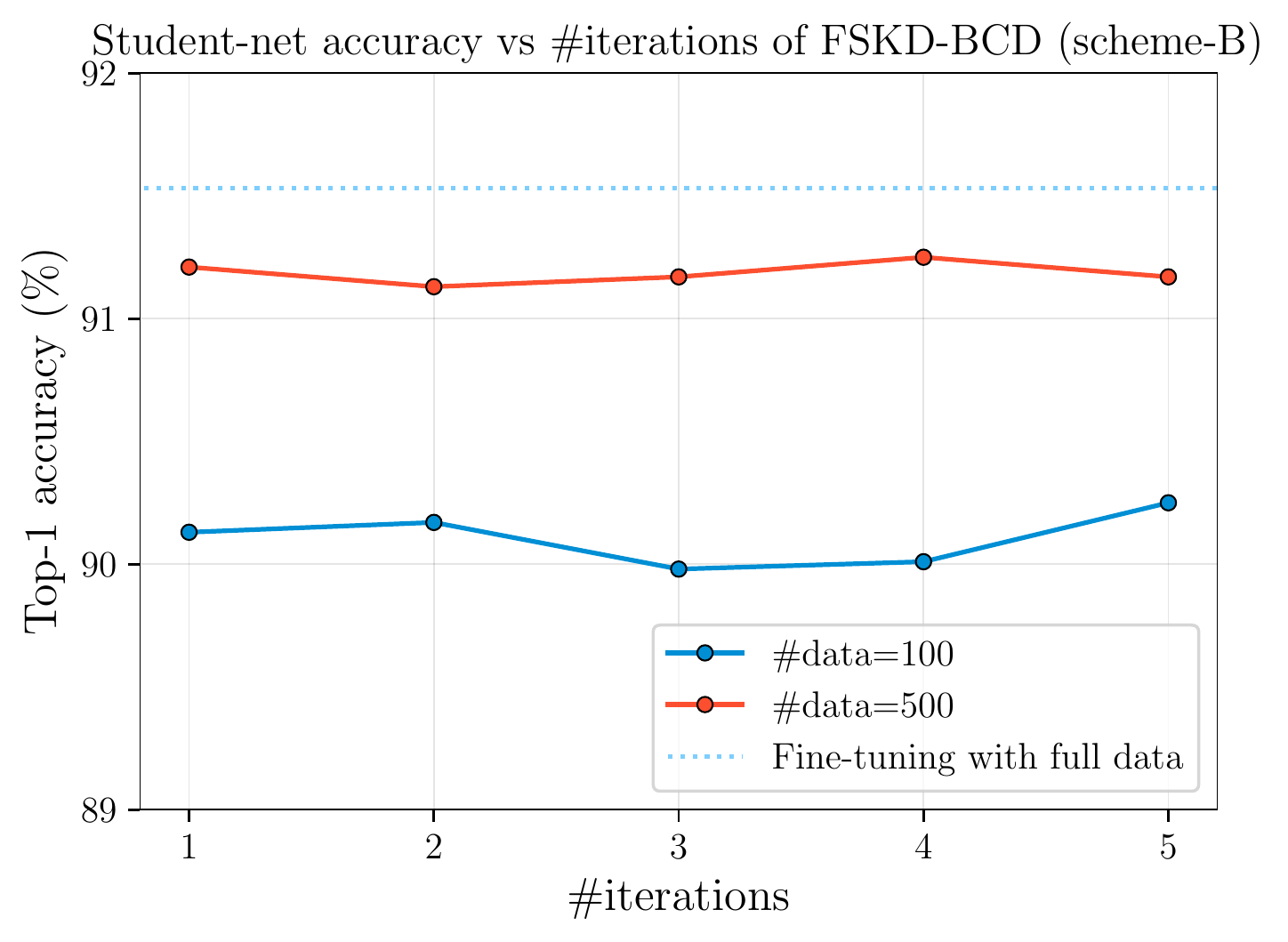}
\vspace{-2.5ex}
\caption{Accuracy vs \#iterations of FSKD on CIFAR-10. Student-net is Prune-B by \textbf{filter pruning}.}
\label{fig:iteration_acc}
\end{figure}

 In our FSKD algorithm, we can apply the block-coordinate descent for several iterations. However, we do not observe noticeable gains for the iteration number $T>1$ over $T=1$ as shown in \autoref{fig:iteration_acc}, so that we set $T=1$ in all our following experiments. 
This may be due to the reason that in each iteration, the sub-problem is a linear optimization problem so that we can find exact minimization, which is consistent with the finding by \cite{hong2017iteration}.

\section*{B: FSKD-BCD vs. FSKD-SGD}

\begin{table}[htbp]
\small
    \centering
    \resizebox{0.99\linewidth}{!}{%
    \begin{tabular}{lccc}
    \hline
    & Acc. (\%) & \#Samples & Time (sec)\\
    \hline
    VGG-16     & 92.66 & 50000 &\\
     Prune-B + FSKD-BCD   & 90.17 & 100 & 3.7\\
     Prune-B + FSKD-SGD & 89.41  & 100 & 18.4 \\
     Prune-B + FSKD-BCD   & {91.21} & 500 & 19.3\\
     Prune-B + FSKD-SGD & 90.76 & 500 & 50.5\\
    \hline
    \end{tabular}
    }
    \caption{Performance comparison between FSKD-BCD and FSKD-SGD by student-nets from \textbf{filter pruning} of VGG-16 with pruning scheme B on CIFAR-10.}
  \label{tab:bcd_sgd}
\end{table}
In this section, we compared two FSKD optimization algorithms: FSKD-BCD uses the BCD algorithm on block-level and FSKD-SGD optimizes the loss all together with the standard SGD algorithm. We valuate both methods on VGG-16 models trained on CIFAR-10 and compressed using filter pruning (prune-B). As shown in  \autoref{tab:bcd_sgd}, FSKD-BCD achieves better accuracy than FSKD-SGD while significantly improves time efficiency.

\section*{C: Proof of Theorem 1}
\begin{proof}
When $\mathbf{W}$ is a point-wise convolution with tensor $\mathbf{W} \in \mathbb{R} ^{n_o\times n_i\times 1 \times 1}$, both $\mathbf{W}$ and $\mathbf{Q}$ are degraded into matrix form. It is obvious that when condition $c1\sim c3$ satisfied, the theorem holds with $\mathbf{W}' = \mathbf{Q} * \mathbf{W}$ in this case, where $*$ indicates matrix multiplication.

When $\mathbf{W}$ is a regular convolution with tensor $\mathbf{W} \in \mathbb{R} ^{n_o\times n_i\times k \times k}$, the proof is non-trivial.
Fortunately, recent work on network decoupling \cite{guo2018nd} presents an important theoretic result as the basis of our derivation.
\begin{lemma}\label{lemma1}
    Regular convolution can be exactly expanded to a sum of several depth-wise separable convolutions.
    Formally, $\forall~ \mathbf{W} \in \mathbb{R} ^{n_o\times n_i\times k \times k}$, $\exists$ $\{\mathbf{P}_k, \mathbf{D}_k\}_{k=1}^K$,
    where $\mathbf{P}_k \in \mathbb{R}^ {n_o\times n_i\times 1\times 1}$, $\mathbf{D}_k\in \mathbb{R} ^{1\times n_i\times k \times k}$,
    \begin{equation}\label{eqndwpw}
    \begin{split}
    s.t.~&~(a) K\le k^2; \\
    &~(b) \mathbf{W} = \sum\nolimits_{k=1}^K \mathbf{P}_k\circ \mathbf{D}_k,
    \end{split}
    \end{equation}
where $\circ$ is the compound operation, which means performing $\mathbf{D}_k$ before $\mathbf{P}_k$.
\end{lemma}
Please refer to \cite{guo2018nd} for the details of proof for this Lemma.
When $\mathbf{W}$ is applied to an input patch $\mathbf{x} \in \mathbb{R}^{n_i \times k\times k}$, we obtain a response vector $\mathbf{y} \in \mathbb{R}^{n_o}$ as
\begin{equation}
    \mathbf{y} = \mathbf{W} \otimes \mathbf{x},
\label{eq:conv}
\end{equation}
where $y_o = \sum_{i=1}^{n_i}{W}_{o, i} \otimes x_i, o\in [n_o], i\in [n_i]$, and $\otimes$ here means convolution operation.
${W}_{o, i} = \mathbf{W}[o, i, :, :]$ is a tensor slice along the $i$-th input and $o$-th output channels,
$x_i = \mathbf{x}[i, :, :]$ is a tensor slice along the $i$-th channel of 3D tensor $\mathbf{x}$.

\begin{figure*}[t]
  \centering
  \small
      \subfloat[]{\includegraphics[width = 0.245\textwidth]{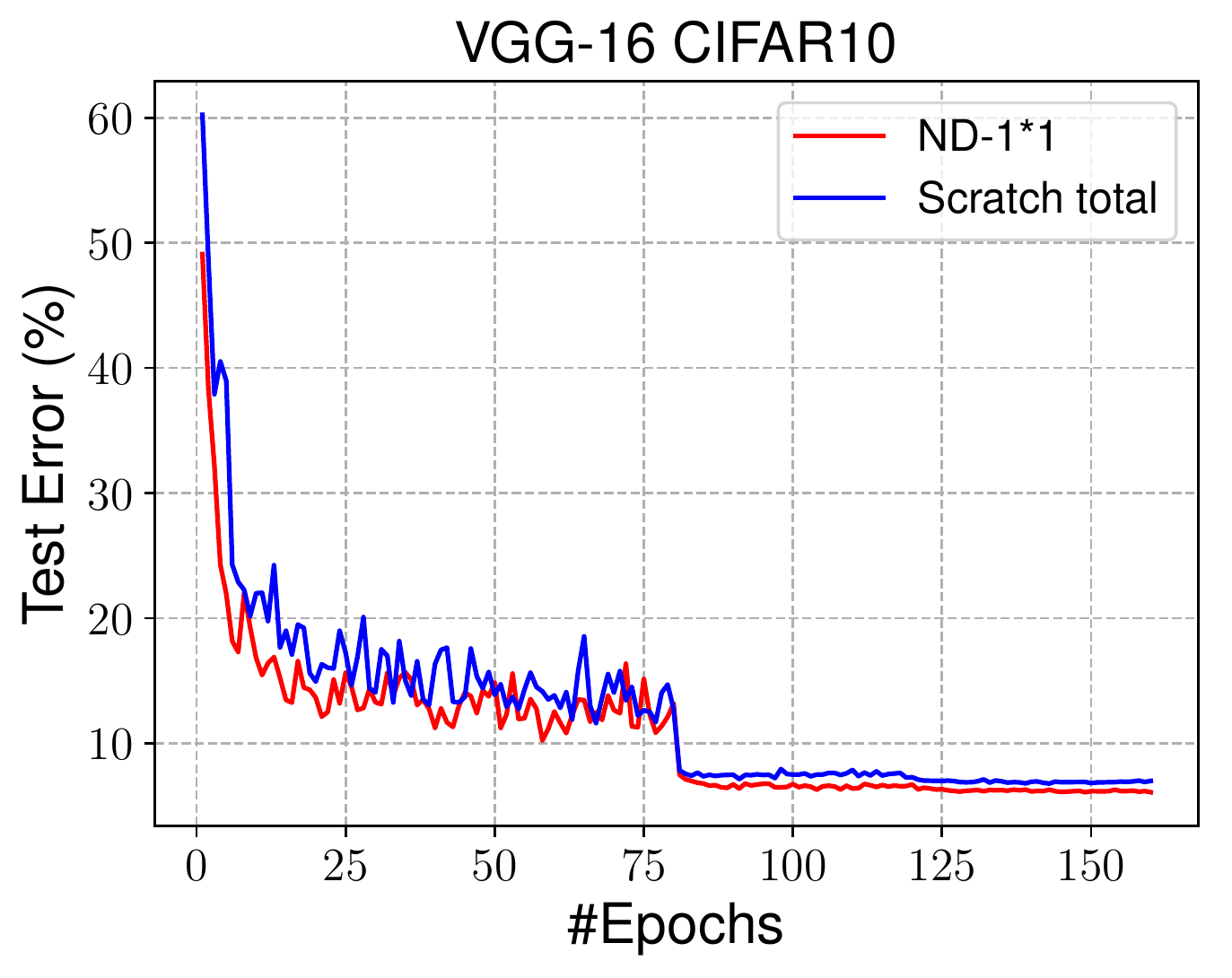}}
      \subfloat[]{\includegraphics[width = 0.245\textwidth]{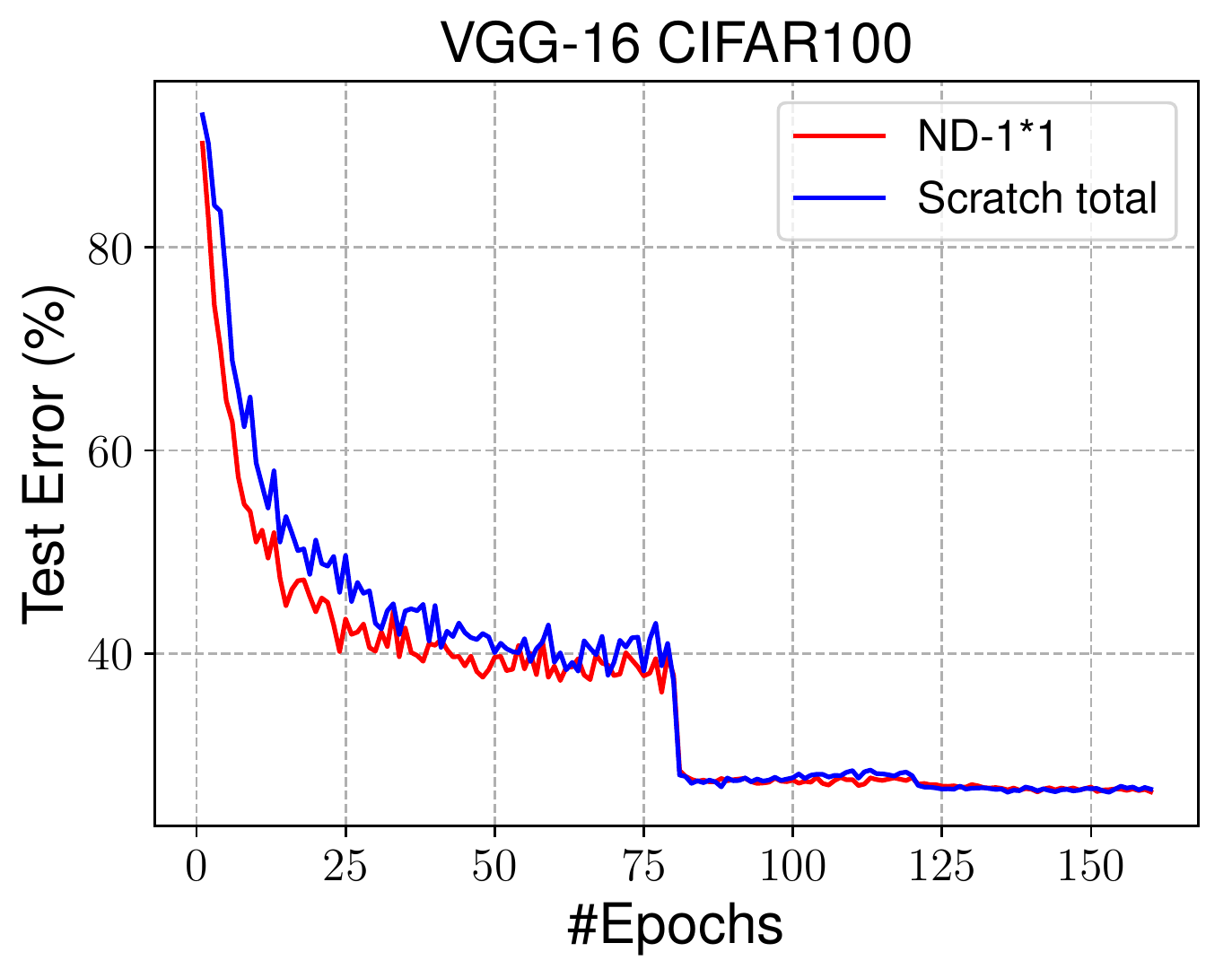}}
      \subfloat[]{\includegraphics[width = 0.245\textwidth]{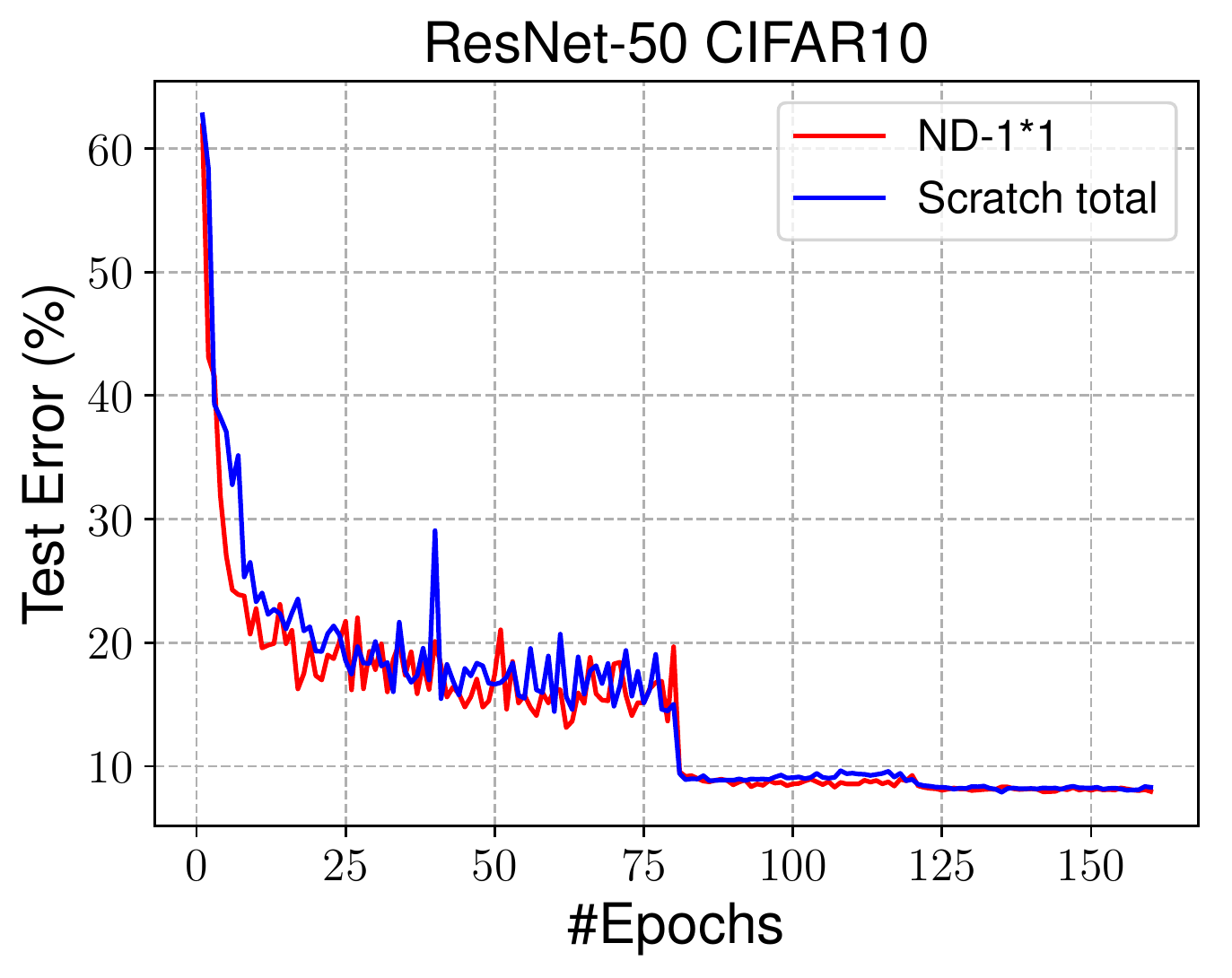}}
      \subfloat[]{\includegraphics[width = 0.245\textwidth]{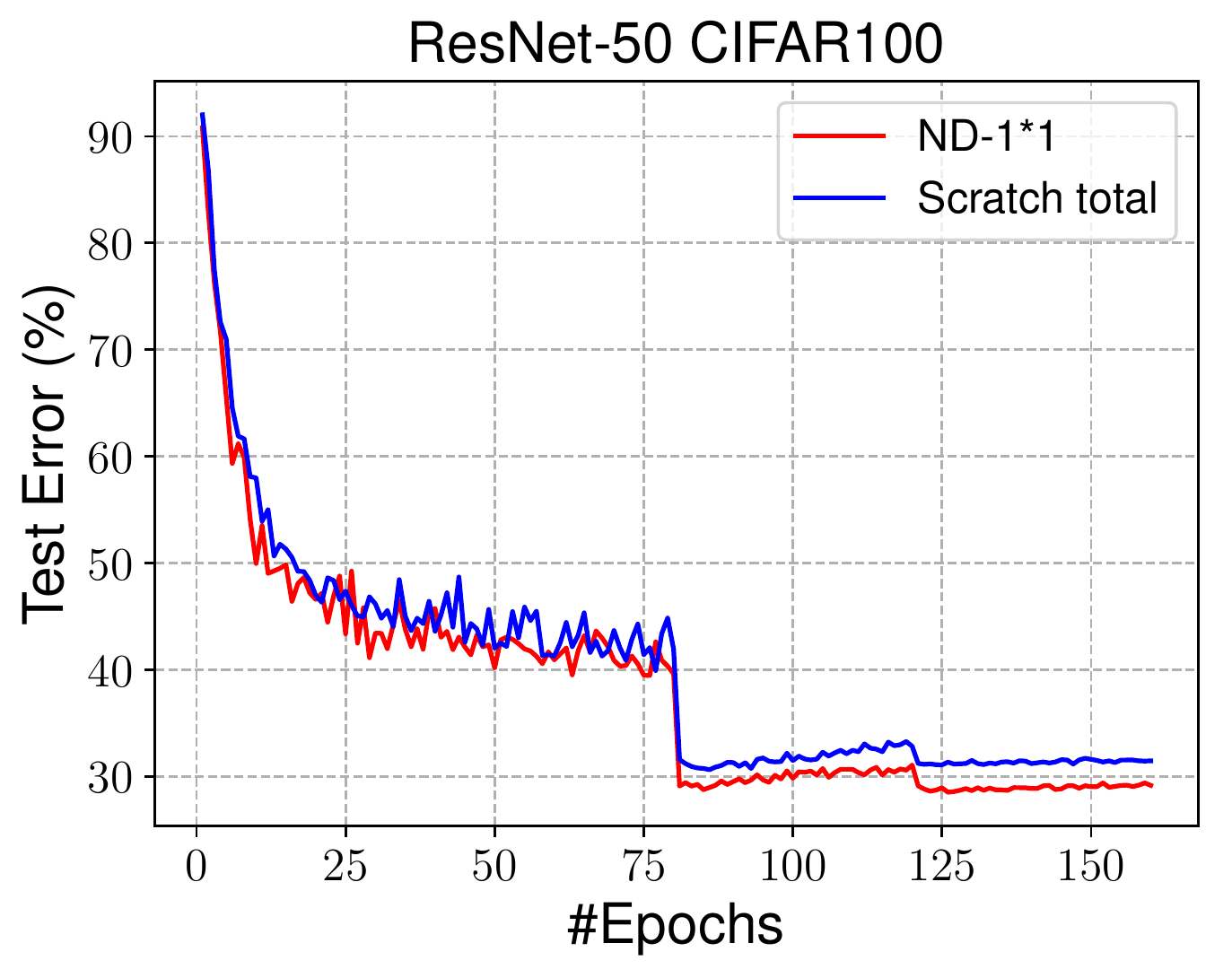}}
      \caption{Test-accuracy at different epochs (a)VGG-16 on CIFAR-10, (b) VGG-16 on CIFAR-100, (c)ResNet-50 on CIFAR-10, (d)ResNet-50 on CIFAR-100.
      ``scratch-total'' is the 1st setting, while ``ND-1*1''is the 2nd setting.
      }
      \label{fig:1x1curve}
  \end{figure*}

When point-wise convolution $\mathbf{Q}$ is added after $\mathbf{Q}$ without non-linear activation between them, we have
\begin{equation}
    \mathbf{y}' = \mathbf{Q} \circ (\mathbf{W} \otimes \mathbf{x}).
\label{eq:conv2}
\end{equation}
With Lemma-1, we have
\begin{equation}
    \mathbf{y}' = (\mathbf{Q}\circ \sum\nolimits_{k=1}^K \mathbf{P}_k\circ \mathbf{D}_k) \otimes \mathbf{x} = (\sum\nolimits_{k=1}^K (\mathbf{Q} * \mathbf{P}_k) \circ \mathbf{D}_k) \otimes  \mathbf{x}
\end{equation}
As both $\mathbf{Q}$ and $\mathbf{P}_k$ are degraded into matrix form, denoting $\mathbf{P}^{'}_k = \mathbf{Q} * \mathbf{P}_k$ and $\mathbf{W'} = \sum\nolimits_{k=1}^K \mathbf{P}^{'}_k \circ \mathbf{D}_k$, we have $\mathbf{y}' = \mathbf{W'}\circ \mathbf{x}.$  This proves the case when $\mathbf{W}$ is a regular convolution.
\end{proof}

\section*{D: Algorithm for iterative pruning and FSKD}
Algorithm-\ref{iterfskd} describes the iteratively pruning and FSKD procedure to achieve extremely compression rate based on \cite{han2015deep,li2016pruning,liu2017learning}.
\begin{algorithm}[h]
\begin{footnotesize}
\caption{Iteratively pruning and FSKD Algorithm}
\label{iterfskd}
\begin{algorithmic}[1]
\INPUT {Teacher-net $\bm{t}$, input data $\{\bm{X}_i\}_{i=1}^N$, \\prune-ratio-list $\{r_k\}_{k=1}^K$, number of iterations $T$}
\STATE $\bm{s}_{max}=\varnothing$\;
\FOR{$t=1:T$}
    \STATE $q_{max} = 0$\;
    \FOR{$k=1:K$}
        \STATE Prune $\bm{s}$ with ratio $r_k$ to obtain student-net $\bm{t}$\;
        \STATE Run FSKD with $\bm{s}$, $\bm{t}$ and $\{\bm{X}_i\}_{i=1}^N$, output $\bm{s}'$\;
        \STATE Evaluation $\bm{s}'$ on validation set to obtain score $q_k$\;
        \IF{$q_k > q_{mqx}$}
            \STATE $q_{max} = q_k$\;
            \STATE $\bm{s}_{max}=\bm{s}'$\;
        \ENDIF
    \ENDFOR
    \STATE Update teacher $\bm{t} = \bm{s}_{max}$\;
\ENDFOR
\OUTPUT {final student-net $\bm{s}_{max}$.}
\end{algorithmic}
\end{footnotesize}
\end{algorithm}
  
\section*{E: Training only PW conv-layer is enough}
People may challenge that learning $1\times 1$-conv may loss representation power and ask why the added $1\times 1$ convolution works so well with such few samples.
According to the network decoupling theory (Lemma-\ref{lemma1}), any regular conv-layer could be decomposed into a sum of depthwise separable blocks, where each depthwise separable block consists of a depthwise (DW) convolution (for spatial correlation modeling) followed by a pointwise (PW) convolution (for cross-channel correlation modeling).
The added $1\times 1$ conv-layer is absorbed/merged into the previous PW layer finally.
The decoupling yields that the number of parameters in PW-layer occupies most ($>$=80\%) parameters of the whole network.
We argue that learning only $1\times 1$-conv is still very powerful, and make a \textbf{bold hypothesis} that PW conv-layer is more critical for performance than DW conv-layer.
To verify this hypothesis, we conduct experiments on VGG16 and ResNet50 on CIFAR-10 and CIFAR-100 under below different settings.
\begin{itemize}
\setlength{\topsep}{1pt}
\setlength{\itemsep}{1pt}
\setlength{\parskip}{1pt}
\item[(1)] We train the network from random initialization with 160 epochs with learning-rate decay 1/10 at 80, 120 epochs from 0.01 to 0.0001.
\item[(2)] We start from a random initialized network (VGG16 or ResNet50), and do full rank decoupling ($K=k^2$ in Eq. \ref{eqndwpw}) so that channels in DW layers are orthogonal, and PW layers are still fully random. Note that Lemma-\ref{lemma1} ensures the network before and after decoupling are equivalent (i.e., able to transfer back and force from each other).
    We keep all the DW-layers fixed (with orthogonal basis from random data), and train only the PW layers with 160 epochs.
    We denote this scheme as {ND-1*1}.
\end{itemize}

\begin{table}[htbp]
  \small
    \centering
    \begin{tabular}{lcc}
    \hline
    Model & CIFAR-10(\%) & CIFAR-100(\%)\\
    \hline
    VGG-16        &  93.00 & 73.35\\
    VGG-16 (ND-1*1) & 93.91 & 73.61 \\
    \hline
    ResNet-50       &   92.64 & 69.93\\
    ResNet-50 (ND-1*1) &  93.51 & 70.83\\
    \hline
    \end{tabular}
    \caption{Results by two schemes (1) full training (2) only training pointwise conv-layers (ND-1*1).}
  \label{tab:1x1}
\end{table}

Note that except the setting explicitly described, all the other configurations (including training epochs, hyper-parameters, hardware platform, etc) are kept the same on both experimental cases.
\autoref{tab:1x1} lists the experimental results on these two cases on both datasets with two different network structures.
It is obvious that the 2nd case (ND-1*1) clearly outperforms the 1st case.
\autoref{fig:1x1curve} further illustrates the test accuracy at different training epochs, which clear shows that the 2nd case (ND-1*1) converges faster and better than the 1st case. This experiment verifies our hypothesis that when keeping DW channels orthogonal, training only the pointwise ($1\times 1$) conv-layer is accurate enough, or even better than training all the parameters together.

\section*{F: Filter Visualization}
\begin{figure}[]
     \centering
     \small
     \includegraphics[width=0.95\linewidth]{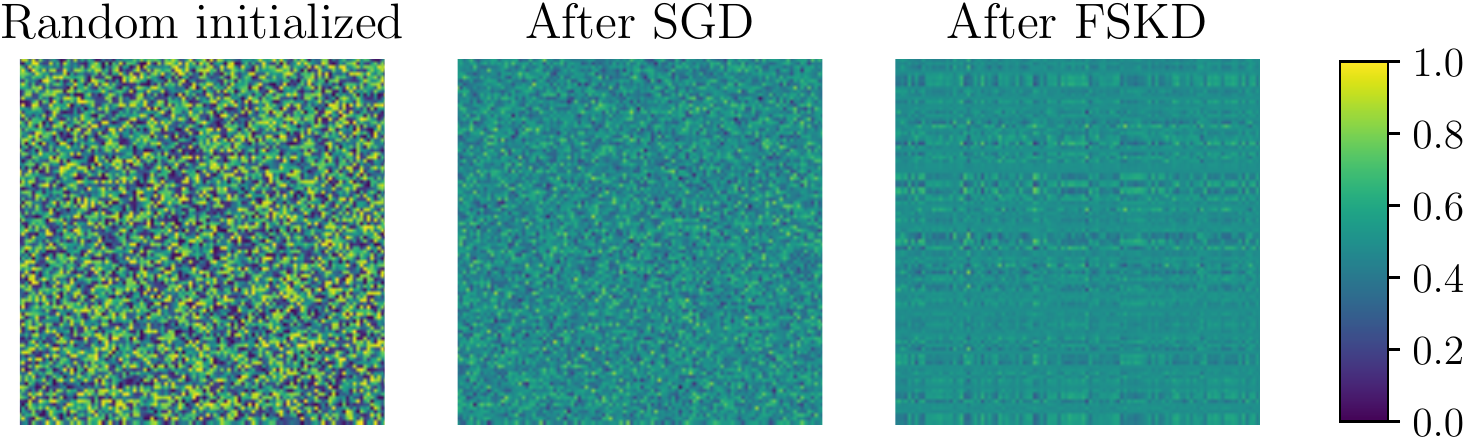}
     \vspace{-2ex}
     \caption{Decouple VGG-13 into DW conv-layer and PW conv-layers, and show one PW conv-layer with random initialization (left), after SGD based fine-tuning (middle), and after FSKD (right). Note values of the PW tensor are scaled into the range (0,1.0) by the min/max values of the tensor for better visualization.}
     \label{fig:filter_viz}
     \vspace{-3ex}
 \end{figure}

 In this section, we try to answer why FSKD works so well that it can provide almost the same results as that of fine-tuning with full training set.
 We conduct experiments based on VGG-13 on CIFAR-10.
 For a given VGG-13 network, We first decouple a conv-layer to obtain one DW conv-layer and one PW conv-layer, as is done in network decoupling \cite{guo2018nd}. Then we visualize the PW conv-layer of the decoupled layer. For simplicity, we only visualize the PW conv-layer of the first decoupled layer. We do the visualization on three VGG-13 network with different parameters:

 \begin{itemize}
     \vspace{-1ex}
     \setlength{\itemsep}{1pt}
     \setlength{\parskip}{1pt}
     \item[(1)] Initialize the VGG-13 network with the MSRA initialization (\autoref{fig:filter_viz} left).
     \item[(2)] Run SGD based fine-tuning on 500 samples for VGG-13 with random initialization until convergence (\autoref{fig:filter_viz} middle).
     \item[(3)] Run FSKD on 500 samples for VGG-13 with SGD based initialization (\autoref{fig:filter_viz} right). The teacher network is also a VGG-13 trained on full CIFAR-10 training set.
     \vspace{-1ex}
 \end{itemize}
 It clearly shows that the PW conv-layer before fine-tuning is fairly random on the value range, the one after fine-tuning is less random, while the one after FSKD further starts to show some regular patterns, which demonstrates that FSKD can distill the knowledge from the teacher-net to student-net effectively with few samples.

\end{document}